\documentclass[10pt,twocolumn,letterpaper]{article}
\PassOptionsToPackage{numbers, sort&compress}{natbib}
\usepackage[pagenumbers]{cvpr} %

\usepackage[utf8]{inputenc} %
\usepackage[T1]{fontenc}    %
\usepackage{url}            %
\usepackage{booktabs}       %
\usepackage{amsfonts}       %
\usepackage{nicefrac}       %
\usepackage{microtype}      %

\usepackage[table,x11names,dvipsnames]{xcolor}
\usepackage{tikz}
\usepackage[framemethod=tikz]{mdframed}
\usepackage{graphicx}
\usepackage[export]{adjustbox} %

\usepackage{mathtools}
\usepackage{amsthm}
\usepackage{stmaryrd}

\usepackage{tabularx}
\usepackage{makecell}

\usepackage{multirow} %
\usepackage{algorithm}
\usepackage{algorithmic}

\usepackage{wrapfig}
\usepackage{comment}
\usepackage{amsmath,amssymb} %
\usepackage{symbols}
\usepackage{pifont} %

\usepackage{amsmath,amsfonts,bm}

\def\floor#1{\lfloor #1 \rfloor}
\def\1{\bm{1}}

\def\sp{space}

\def\ve{{\bm{e}}}

\def\vh{{\bm{h}}}

\def\vr{{\bm{r}}}

\def\vt{{\bm{t}}}

\def\vv{{\bm{v}}}

\def\vx{{\bm{x}}}
\def\vy{{\bm{y}}}
\def\vz{{\bm{z}}}

\def\mB{{\bm{B}}}

\def\mE{{\bm{E}}}

\def\mH{{\bm{H}}}
\def\mI{{\bm{I}}}

\def\mK{{\bm{K}}}

\def\mQ{{\bm{Q}}}

\def\mT{{\bm{T}}}

\def\mV{{\bm{V}}}

\def\mX{{\bm{X}}}
\def\mY{{\bm{Y}}}
\def\mZ{{\bm{Z}}}

\DeclareMathAlphabet{\mathsfit}{\encodingdefault}{\sfdefault}{m}{sl}
\SetMathAlphabet{\mathsfit}{bold}{\encodingdefault}{\sfdefault}{bx}{n}

\def\gD{{\mathcal{D}}}

\def\gS{{\mathcal{S}}}

\def\sR{{\mathbb{R}}}

\def\sZ{{\mathbb{Z}}}

\newcommand{\circshift}[2]{#1\ {\tt mod}\ #2}

\DeclareMathOperator*{\argmax}{arg\,max}

\newcommand*{\ShowNotes}{} %
\definecolor{darkred}{rgb}{0.7,0.1,0.1}
\definecolor{darkgreen}{rgb}{0.1,0.7,0.1}
\definecolor{cyan}{rgb}{0.7,0.0,0.7}
\definecolor{dblue}{rgb}{0.2,0.2,0.8}
\definecolor{maroon}{rgb}{0.76,.13,.28}
\definecolor{burntorange}{rgb}{0.81,.33,0}
\definecolor{tealblue}{rgb}{0.212,0.459, 0.533}

\definecolor{mypink}{rgb}{0.93359375, 0.62109375, 0.83984375}

\definecolor{pp}{rgb}{0.43921569, 0.18823529, 0.62745098}
\definecolor{rr}{rgb}{0.5254902 , 0.00784314, 0.12941176}
\definecolor{bb}{rgb}{0.09019608, 0.23529412, 0.37647059}
\definecolor{yy}{rgb}{0.49803922, 0.3372549 , 0.0}
\definecolor{gg}{rgb}{0.02352941, 0.3372549 , 0.17647059}

\ifdefined\ShowNotes
  \newcommand{\colornote}[3]{{\color{#1}\bf{#2: #3}\normalfont}}
\else
  \newcommand{\colornote}[3]{}
\fi

\newcommand{\eat}[1]{} %

\usepackage{pmboxdraw}
\usepackage[multiple]{footmisc}

\newcommand{\myparagraph}[1]{\vspace*{0pt}{\bf\noindent #1}}

\usepackage{cancel}

\definecolor{mybrown}{rgb}{0.87058824, 0.56078431, 0.01960784}
\definecolor{myblue}{rgb}{0.3372549 , 0.70588235, 0.91372549}
\definecolor{mypurple}{rgb}{0.8, 0.47058824, 0.7372549 }
\definecolor{myorange}{rgb}{0.835, 0.368, 0}
\definecolor{mygreen}{rgb}{0.00784314, 0.61960784, 0.45098039}
\definecolor{mygt}{rgb}{0.0078125 , 0.57421875, 0.40625}
\definecolor{mysp}{rgb}{0.84765625, 0.515625  , 0.0234375}
\definecolor{mycitecolor}{rgb}{0,0.08,0.45}
\definecolor{mygr}{rgb}{0.9607,0.9607,0.9607}
\definecolor{myoo}{rgb}{0.992,0.9176,0.9019}
\definecolor{myrr}{HTML}{AE031A}
\definecolor{mybb}{HTML}{0155B3}
\definecolor{bg_blue}{HTML}{E4EFFF}

\usepackage{thm-restate}
\mdfsetup{skipabove=-1.5cm,skipbelow=-1.5cm}
\mdfdefinestyle{MyFrame}{%
    linecolor=Black,
    outerlinewidth=0.1pt,
    roundcorner=2pt,
    innertopmargin=-.45\baselineskip,
    innerbottommargin=0.3\baselineskip,
    innermargin = 0.cm,
    outermargin = 0.cm,
    innerrightmargin=5pt,
    innerleftmargin=5pt,
    backgroundcolor=black!2!white}
    
\mdfdefinestyle{s02}{%
    linecolor=Black,
    outerlinewidth=0.1pt,
    roundcorner=2pt,
    innertopmargin=0.4\baselineskip,
    innerbottommargin=0.4\baselineskip,
    innermargin = 0.2cm,
    outermargin = 0.cm,
    backgroundcolor=Green!0!white}

\usepackage[textsize=tiny]{todonotes}
\usepackage[frozencache]{minted}
\usepackage{listings}

\definecolor{cvprblue}{rgb}{0.21,0.49,0.74}
\definecolor{lightcarminepink}{rgb}{0.9, 0.4, 0.38}
\usepackage[pagebackref,breaklinks,colorlinks,citecolor=cvprblue,linkcolor=lightcarminepink]{hyperref}

\def\paperID{3354}
\def\confName{CVPR}
\def\confYear{2024}

\title{Making Vision Transformers Truly Shift-Equivariant}
\author{
Renan A. Rojas-Gomez\textsuperscript{$\star$}\quad\quad Teck-Yian Lim\textsuperscript{$\star$}\quad\quad Minh N. Do\textsuperscript{$\star$}\quad\quad Raymond A. Yeh\textsuperscript{$\dagger$}\\
\textsuperscript{$\star$}University of Illinois at Urbana-Champaign\\
\textsuperscript{$\dagger$}Purdue University\\
{\tt\small \{renanar2, tlim11, minhdo\}@illinois.edu\quad rayyeh@purdue.edu}
}

\begin{document}
\maketitle
\begin{abstract}
For computer vision, Vision Transformers (ViTs) have become one of the go-to deep net architectures. Despite being inspired by Convolutional Neural Networks (CNNs), ViTs' output remains sensitive to small spatial shifts in the input, i.e., not shift invariant. To address this shortcoming, we introduce novel data-adaptive designs for each of the modules in ViTs, such as tokenization, self-attention, patch merging, and positional encoding. With our proposed modules, we achieve true shift-equivariance on four well-established ViTs, namely, Swin, SwinV2, CvT, and MViTv2. Empirically, we evaluate the proposed adaptive models on image classification and semantic segmentation tasks. These models achieve competitive performance across three different datasets while maintaining 100\% shift consistency.
\end{abstract}

\section{Introduction}
Vision Transformers (ViTs)~\cite{dosovitskiy2020vit,liu_2021_swin,wu2021cvt,liu2022swin,fan2021multiscale,li2022mvitv2} have become a strong alternative to convolutional neural networks (CNNs) in computer vision, superseding their dominance in image classification and becoming the state-of-the-art model on ImageNet~\cite{deng2009imagenet}. 
Unlike the original Transformer~\cite{vaswani_2017_attention}  proposed for natural language processing (NLP), ViTs incorporate suitable inductive biases for computer vision. Consider image classification, where an input shift does not change the underlying image label, \ie, the task is shift-invariant. 

Several ViTs accredited shift-invariance as the motivation for the proposed architecture. For instance,~\citet{wu2021cvt} state that their ViT model brings ``desirable properties of CNNs to the ViT architecture (i.e. \textit{shift}, scale, and distortion invariance).'' Similarly,~\citet{liu_2021_swin} found that ``inductive bias that encourages certain translation invariance is still preferable for general-purpose visual modeling.'' Surprisingly, despite these proposed architecture changes, ViTs remain sensitive to spatial shifts. This motivates us to study how to make ViTs truly shift-invariant and equivariant.

In this work, we carefully re-examine each of the building blocks in ViTs and propose novel shift-equivariant versions of each module. This includes redesigning the tokenization, self-attention, patch merging, and positional encoding modules. Our design principle is to perform an alignment that is dependent on the input signal, \ie, each module's behavior is \textit{adaptive} to the input, hence we prepend each module's name with \textit{(A)daptive}. All our adaptive modules are provably shift-equivariant and realizable in practice.

Our adaptive design enables truly shift-invariant ViTs for image classification and truly shift-equivariant ViTs for semantic segmentation,~\ie, achieving $100\%$ shift consistency. We conduct extensive experiments on CIFAR10/100~\cite{krizhevsky2009learning}, and ImageNet for classification; as well as ADE20K~\cite{zhou2017scene} for semantic segmentation. We empirically show that our architectures improve shift consistency and have competitive performance on four well-established ViTs: Swin~\cite{liu_2021_swin}, SwinV2~\cite{liu2022swin},
CvT~\cite{wu2021cvt},
and MViTv2~\cite{li2022mvitv2}.

\vspace{2pt}
\myparagraph{Our contributions are as follows}:
\begin{itemize}
    \item We propose a family of ViT modules that are provably circularly shift-equivariant.
    \item With these proposed modules, we build truly shift-invariant and equivariant ViTs, achieving 100\% circular shift-consistency measured from end-to-end.
    \item Extensive experiments on image classification and semantic segmentation demonstrate the effectiveness of our approach in terms of performance and shift consistency.
\end{itemize}

\begin{figure*}[t]
    \hspace{-0.038\linewidth}
    \subfloat[Original tokenization (\texttt{token}).]{
    \includegraphics[height=0.26\linewidth,valign=t]{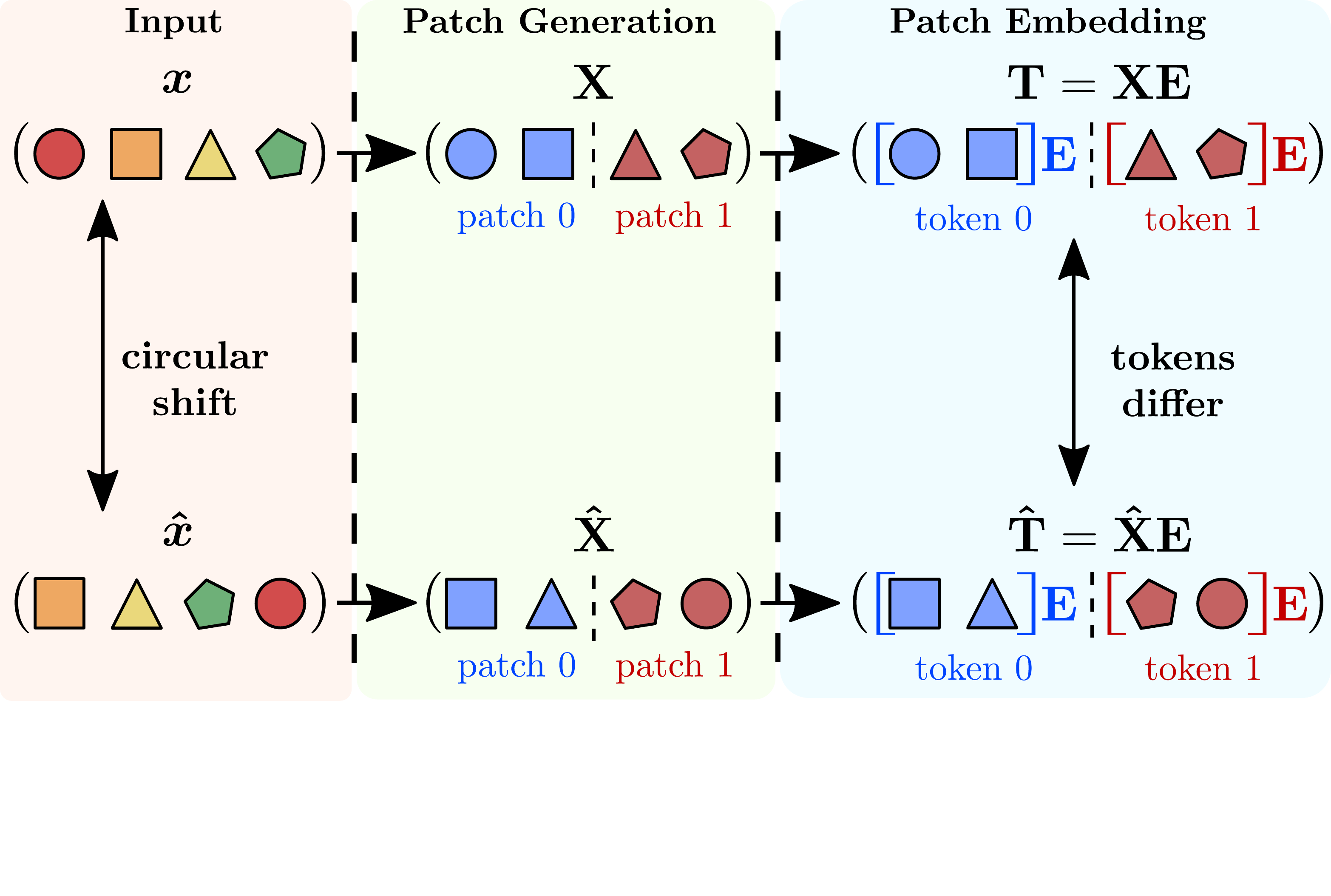}\label{fig:token_orig}\hspace{0.025\linewidth}}
    \subfloat[Proposed adaptive tokenization (\texttt{A-token}).]{\includegraphics[height=0.26\linewidth,valign=t]{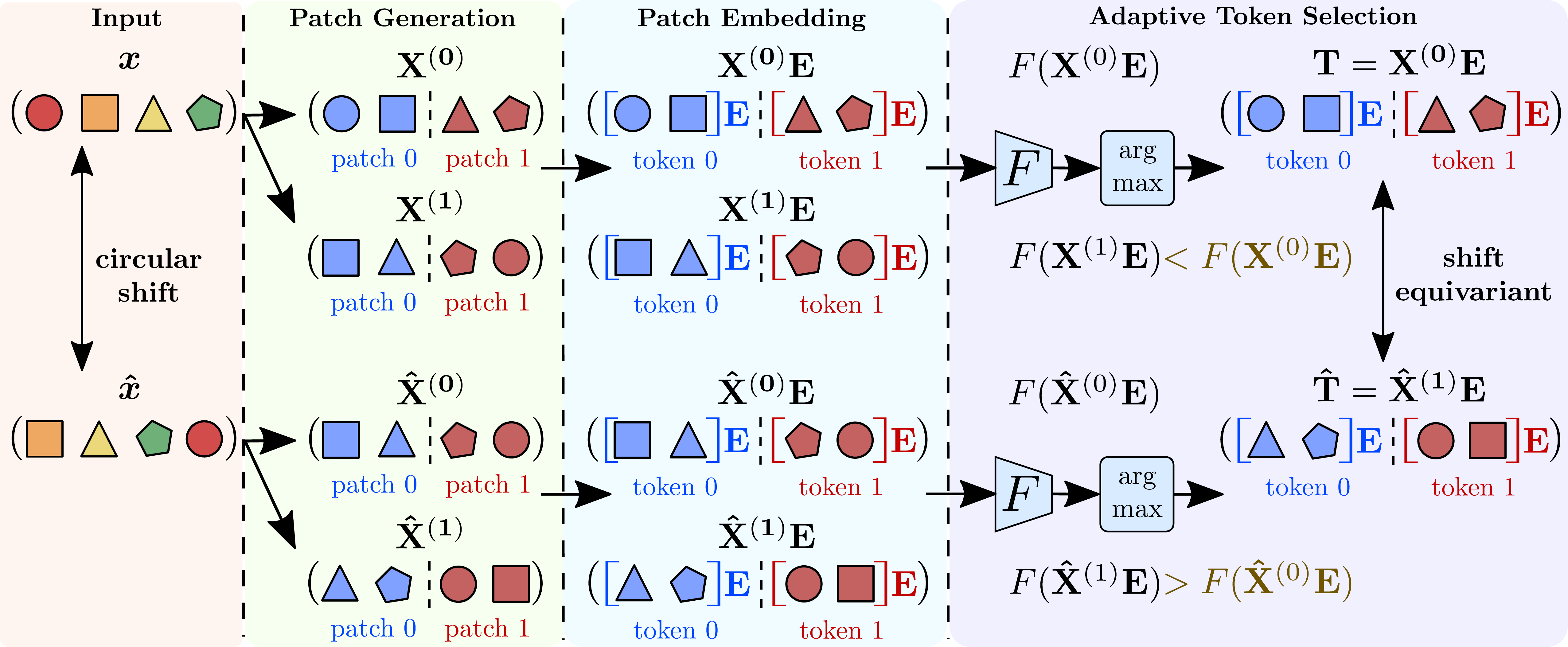}\label{fig:atoken}}
    \vspace{-0.1cm}
    \caption{\textbf{Re-designing ViT's tokenization towards shift-equivariance:} (a) The original patch embedding is sensitive to small input shifts due to the fixed grid used to split an image into patches. (b) Our adaptive tokenization \texttt{A-token} is a generalization that consistently selects the group of patches with the highest energy, despite input shifts.}
    \vspace{-.3cm}
    \label{fig:tokenization}
\end{figure*}
\vspace{-.3cm}

\section{Related Work}
We briefly discuss ViTs and shift-equivariant CNNs. Additional background concepts are reviewed in~\secref{sec:prelim}.

\myparagraph{Vision transformers.}
Proposed by~\citet{vaswani_2017_attention} for NLP tasks, the Transformer architecture incorporates a tokenizer, positional encoding, and attention mechanism into one architecture. Soon after, the Transformer found its success in computer vision by incorporating suitable inductive biases, such as shift equivariance, giving rise to the area of Vision Transformers. Seminal works include: ViT~\cite{dosovitskiy2020vit}, which treats an image as $16\times16$ tokens;  
Swin~\cite{liu_2021_swin,liu2022swin}, which introduces locally shifted windows to the attention mechanism; CvT~\cite{wu2021cvt}, which introduces convolutional layers to ViTs; and MViT~\cite{fan2021multiscale,li2022mvitv2}, which introduces a multi-scale pyramid structure. A more in-depth discussion on ViTs can be found in recent surveys~\cite{han2022survey,khan2022transformers}. In this work, we re-examine ViTs' modules and present a novel adaptive design that enables truly shift-equivariant ViTs.

Concurrently on \textit{ArXiv},~\citet{ding_2023_reviving} propose a polyphase anchoring method to obtain circular shift invariant ViTs in image classification. Differently, our method demonstrates improvements in circular and linear shift consistency for both image classification and semantic segmentation while maintaining competitive task performance.

\myparagraph{Invariant and equivariant CNNs.}
Prior works~\cite{Azulay_Weiss_2019,zhang2019making} have shown that modern CNNs~\cite{Krizhevsky_Sutskever_Hinton_2012,Simonyan_Zisserman_2015,He_Zhang_Ren_Sun_2016,Sandler_Howard_Zhu_Zhmoginov_Chen_2018} are not shift-equivariant due to the usage of pooling layers. To improve shift-equivariance, anti-aliasing~\cite{vetterli2014foundations} is introduced before downsampling. Specifically,~\citet{zhang2019making} and ~\citet{zou2020delving} propose to use a low-pass filter (LPF) for anti-aliasing.

While anti-aliasing improves shift-equivariance, the overall CNN remains not truly shift-equivariant. To address this,~\citet{chaman2021truly} propose Adaptive Polyphase Sampling (APS), which selects the downsampling indices based on the $\ell_2$ norm of the input's polyphase components.~\citet{rojas-neurips2022-learnable} then improve APS by proposing a learnable downsampling module (LPS) that is truly shift-equivariant.~\citet{michaeli_2023_alias} propose the use of polynomial activations to obtain shift-invariant CNNs. In contrast, we present a family of modules that enables truly shift-equivariant ViTs. We emphasize that CNN methods are \textit{not applicable} to ViTs due to their distinct architectures.

Beyond the scope of this work, general equivariance~\cite{cohen2016group,bronstein2017geometric,ravanbakhsh2017equivariance,weiler2019general,venkataraman2019building,romero2020attentive,yeh2022equivariance,shakerinava21a,van2022relaxing,romero2022learning,klee2023image} have also been studied. Equivariant networks have also been applied to sets~\cite{ravanbakhsh_sets,zaheer2017deep, qi2017pointnet, hartford2018deep,yeh2019chirality,maron2020learning}, graphs~\cite{shuman2013emerging,defferrard2016convolutional,kipf2017semi,maron2018invariant,yeh2019diverse,dehaan2020gauge,liu2020pic,liu2021semantic,morris22a}, spherical images~\cite{cohen2018spherical,kondor2018clebsch,cohen2019gauge},~\etc.

\begin{figure*}[t]
    \centering
    \subfloat[Window-based self-attention~(\texttt{WSA})]{
    \hspace{-0.035\linewidth}
    \includegraphics[height=0.225\linewidth]{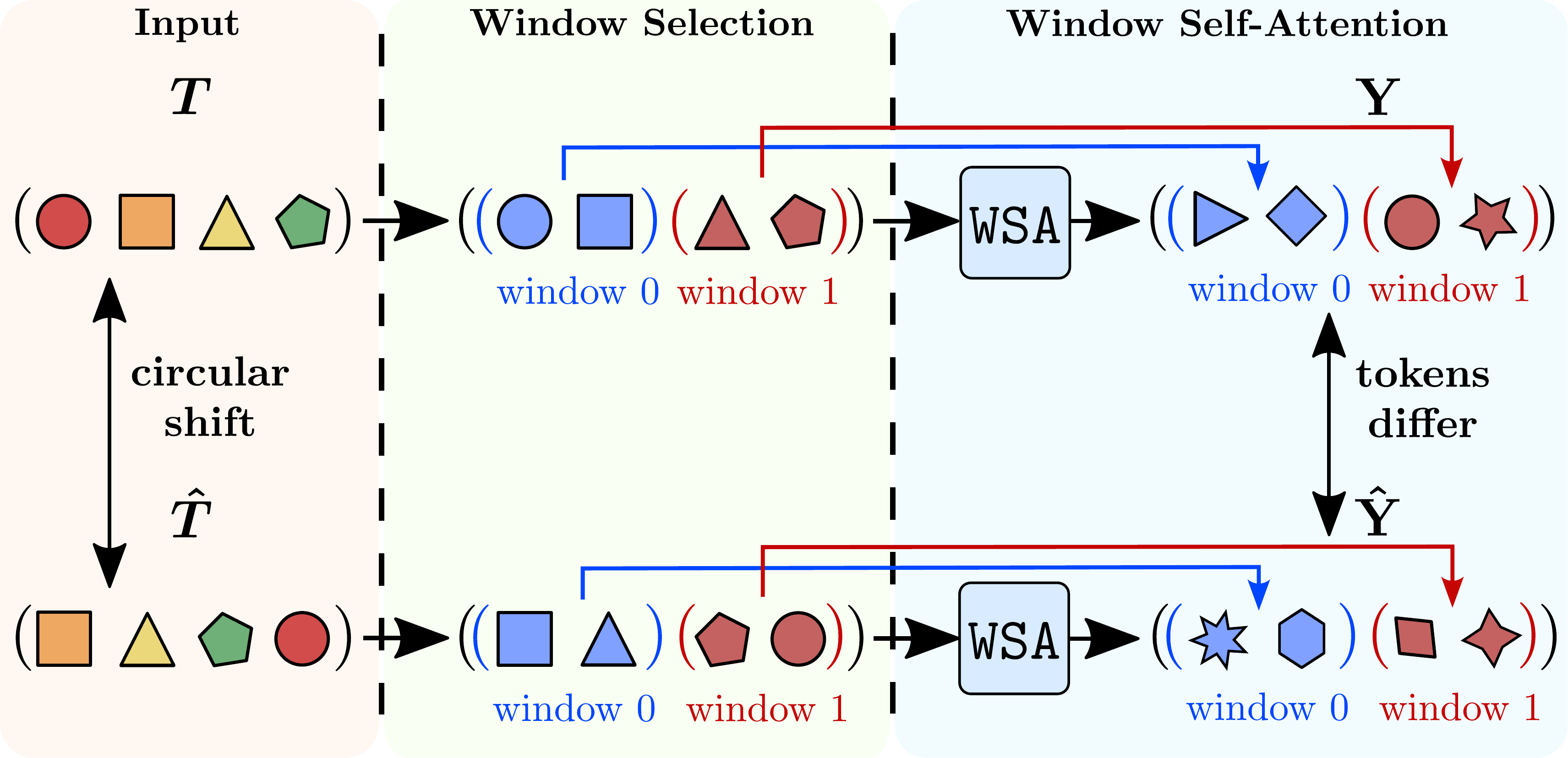}\label{fig:wsa_orig}\hspace{0.025\linewidth}}
    \subfloat[Proposed adaptive window-based self-attention~(\texttt{A-WSA})]{\includegraphics[height=0.225\linewidth]{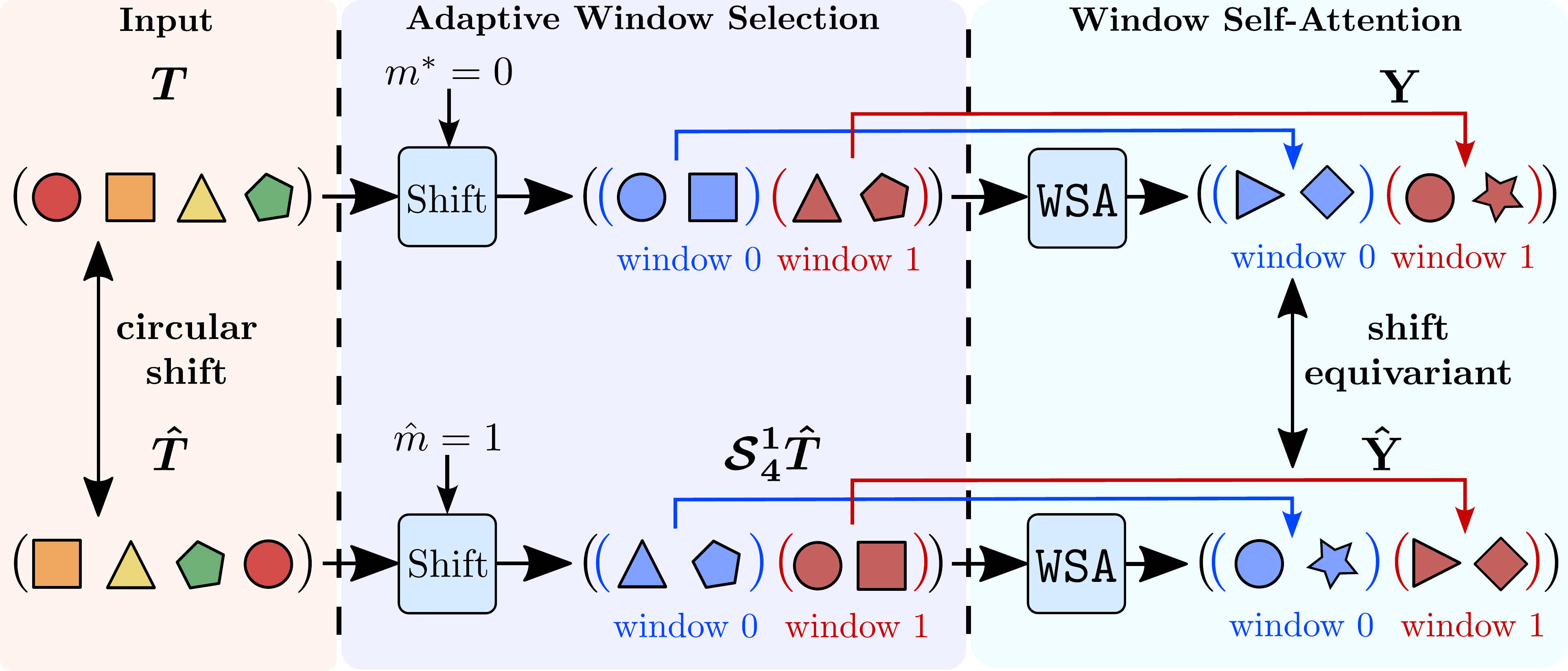}\label{fig:awsa}}
    \vspace{-0.1cm}
    \caption{\textbf{Re-designing window-based self-attention towards shift-equivariance:} (a) The window-based self-attention \texttt{WSA} breaks shift equivariance by selecting windows without considering their input properties. (b) Our proposed adaptive window-based self-attention selects the best grid of windows based on their average energy, obtaining windows comprised of the same tokens despite input shifts.}
    \label{fig:wsa}
    \vspace{-0.2cm}
\end{figure*}

\section{Preliminaries}\label{sec:prelim}
We review the basics before introducing our approach, including the aspects of current ViTs that break shift-equivariance. For readability, the concepts are described in 1D. In practice, these are extended to multi-channel images.

\myparagraph{Equivariance.}
Conceptually, equivariance describes how a function's input and output are related to each other under \textit{a predefined set of transformations}. For example, in image segmentation, shift equivariance means that \textit{shifting} the input image results in \textit{shifting} the output mask. In our analysis, we consider \textit{circular shifts} and denote them as
\bea
\big(\gS_N\vx\big)[n] = \vx[(n+1)~{\tt mod}~N], \vx \in \sR^{N}
\eea 
to ensure that the shifted signal $\vx$ remains within its support.
Following~\citet{rojas-neurips2022-learnable}, we say a function $f: \sR^N \mapsto \sR^M$ is $\gS_N,\{\gS_M, \mI\}$-equivariant, \ie, shift-equivariant, iff $\exists~\gS \in \{\gS_M, \mI\}$ s.t.
\begin{align}\label{eq:def_shift_eq}
 f(\gS_N\vx) =&\ \gS f(\vx)~\forall \vx \in \sR^{N}
\end{align}
where $\mI$ denotes the identity mapping.
This definition carefully handles the case where $N>M$. For instance, when downsampling by a factor of two, an input shift by one should ideally induce an output shift by $0.5$, which is not realizable on the integer grid. This $0.5$ has to be rounded up or down, hence a shift $\gS_M$ or a no-shift $\mI$, respectively.

\myparagraph{Invariance.} For classification, a label remains \textit{unchanged} when the image is \textit{shifted}, \ie, it is shift-invariant. A function $f: \sR^N \mapsto \sR^M$ is $\gS_N,\{\mI\}$-equivariant (invariant) iff
\begin{align}
\label{eq:def_shift_in}
f(\gS_N\vx) =  f(\vx)~\forall \vx \in \sR^{N}.
\end{align}
A common way to design a shift-invariant function under circular shifts is via \textit{global spatial pooling}~\cite{lin2013network}, defined as $g(\vx)=\sum_{m}\vx[m]$. Given a shift-equivariant function $f$:
\bea
\sum_m f(\gS_N\vx)[m] = \sum_m \gS f(\vx)[m] = \sum_m f(\vx)[m].
\eea
However, note that ViTs using global spatial pooling after extracting features are \textit{not shift-invariant}, as preceding layers such as tokenization, window-based self-attention, and patch merging are not shift-equivariant, which we review next.

\myparagraph{Tokenization (\texttt{token}).} 
ViTs split an input $\vx \in \sR^N$ into non-overlapping patches of length $L$ and project them into a latent space to generate tokens.
\begin{align}
    \label{eq:token}
    \text{\texttt{token}}(\vx)=&\ \mX\mE \in \sR^{\frac{N}{L}\times D}
\end{align}
where $\mE\in \sR^{L\times D}$ is a linear projection and $\mX=\texttt{reshape}(\vx) = \begin{bmatrix}\mX_{0}\ & \dots\ & \mX_{\frac{N}{L}-1}\end{bmatrix}^{\top}\in \sR^{\frac{N}{L}\times L}$ is a matrix where each row corresponds to the $k^{\text{th}}$ patch of $\vx$, \ie,
\begin{align}
    \mX_{k}=\vx[Lk:L(k+1)-1]\in \sR^{L}.
\end{align}
\equref{eq:token} implies that \texttt{token} is not shift-equivariant. Patches are extracted based on a \textit{fixed} grid, so different patches will be obtained if the input is shifted, as illustrated in~\figref{fig:token_orig}.

\myparagraph{Self-Attention (\texttt{SA}).} 
In ViTs, self-attention is defined as
\begin{align}
    \label{eq:self_attention}
    \text{\texttt{SA}}(\mT)=&\ \text{\texttt{softmax}}(\mQ\mK^{\top}/\sqrt{D'})\mV \in \sR^{M\times D'}
\end{align}
where $\mT=\begin{bmatrix}\mT_{0} & \dots & \mT_{M-1}\end{bmatrix}^{\top}\in \sR^{M\times D}$ denotes input tokens, and \texttt{softmax} is the softmax normalization along rows. Queries $\mQ$, keys $\mK$ and values $\mV$ correspond to:
\begin{align}
    \mQ =&\ \mT \mE^{Q},\ \mK =\ \mT \mE^{K},\ \mV =\ \mT \mE^{V}
\end{align}
with linear projections $\mE^{Q/K/V} \in \sR^{D \times D'}$. The term $\text{\texttt{softmax}}\big(\mQ\mK^{\top}/\sqrt{D'}\big)\in [0,1]^{M\times M}$ ensures that the output token is a convex combination of the computed values.

\myparagraph{Window-based self-attention (\texttt{WSA}).} 
A crucial limitation of self-attention is its quadratic computational cost with respect to the number of input tokens $M$. To alleviate this, window-based self-attention~\citep{liu_2021_swin} 
groups tokens into local windows and then performs self-attention \textit{within} each window. Given input tokens $\mT \in \sR^{M\times D}$ and a window size $W$, window-based self-attention $\text{\texttt{WSA}}(\mT)\in \sR^{M\times D'}$
is defined as:
\begin{align}
    \label{eq:wsa_def}
    \text{\texttt{WSA}}(\mT)=&\ \begin{bmatrix}\text{\texttt{SA}}\big(\bar{\mT}_{W}^{(0)}\big)\ ;\ \dots\ ;\ \text{\texttt{SA}}\big(\bar{\mT}_{W}^{(\frac{M}{W}-1)}\big)\end{bmatrix}
\end{align}
where $\bar{\mT}_{W}^{(k)}=\begin{bmatrix}\mT_{Wk} & \dots & \mT_{W(k+1)-1}\end{bmatrix}^{\top}\in \sR^{W\times D}$ denotes the $k^{\text{th}}$ window comprised by $W$ neighboring tokens ($W$ consecutive rows of $\mT$). Note that Eq. \eqref{eq:wsa_def} uses semicolons ($;$) as row separators.

Swin~\cite{liu_2021_swin, liu2022swin} architectures take advantage of \texttt{WSA} to decrease the computational cost while adopting a shifting scheme (at the window level) to allow long-range connections. We note that \texttt{WSA} is not shift-equivariant, \eg, any shift that is not a multiple of the window size changes the tokens within each window, as illustrated in \figref{fig:wsa_orig}.

\myparagraph{Patch merging (\texttt{PMerge}).}
Given input tokens $\mT=\begin{bmatrix}\mT_{0} & \dots & \mT_{M-1}\end{bmatrix}^{\top}\in \sR^{M\times D}$ and a patch length $P$, patch merging is defined as a linear projection of vectorized token patches:
\begin{align}
    \label{eq_patch_merge01}
    \text{\texttt{PMerge}}(\mT)=&\ \tilde{\mT}\tilde{\mE} \in \sR^{\frac{M}{P}\times \tilde{D}}\\
    \notag \text{with }\tilde{\mT}=& \begin{bmatrix} \text{\texttt{vec}}(\bar{\mT}_{P}^{(0)}) \ \dots \ \text{\texttt{vec}}(\bar{\mT}_{P}^{(\frac{M}{P}-1)})\end{bmatrix}^{\top}.
\end{align}
Here, $\text{\texttt{vec}}(\bar{\mT}_{P}^{(k)})\in \sR^{PD}$ is the vectorized version of the $k^{\text{th}}$ patch $\bar{\mT}_{P}^{(k)}=\begin{bmatrix}\mT_{Pk} & \dots & \mT_{P(k+1)-1}\end{bmatrix}^{\top}\in \sR^{P\times D}$, and $\tilde{\mE} \in \sR^{PD \times \tilde{D}}$ is a linear projection.

\texttt{PMerge} reduces the number of tokens after applying self-attention while increasing the token length, \ie, $\tilde{D}>D$. This is similar to the CNN strategy of increasing the number of channels using convolutional layers while decreasing their spatial resolution via pooling. Since patches are selected on a fixed grid, \texttt{PMerge} is not shift-equivariant.

\myparagraph{Relative position embedding (RPE).} 
As self-attention is permutation equivariant, spatial information must be explicitly added to the tokens. Typically, RPE adds a position matrix representing the relative distance between queries and keys into self-attention as follows:
\begin{align}
    \text{\texttt{SA}}^{(\text{rel})}(\mT)=&\ \text{\texttt{softmax}}\bigg(\frac{\mQ\mK^{\top}}{\sqrt{D'}}+\mE^{(\text{rel})}\bigg)\mV\\
    \text{with } \mE^{(\text{rel})}[i,j]=&\ \mB^{(\text{rel})}[p^{(Q)}_{i}-p^{(K)}_{j}].
\end{align}
Here, $\mE^{(\text{rel})}\in \sR^{M\times M}$ is constructed from an embedding lookup table $\mB^{(\text{rel})} \in \sR^{2M-1}$ and the index $[p^{(Q)}_{i}-p^{(K)}_{j}]$ denotes the distance between the $i^{\text{th}}$ query token at position $p^{(Q)}_{i}$ and the $j^{\text{th}}$ key token at position $p^{(K)}_{j}$. As embeddings are selected based on the relative distance, RPE allows ViTs to capture spatial relationships,~\eg, knowing whether two tokens are spatially nearby.

\section{Truly Shift-equivariant ViT}
\label{sec:app}
In the previous section, we identified components of ViTs that are not shift-equivariant. To achieve shift-equivariance in ViTs, we redesign four modules: tokenization, self-attention, patch merging, and positional embedding. As equivariance is preserved under compositions, ViTs using these modules are end-to-end shift-equivariant.

\myparagraph{Adaptive tokenization (\texttt{A-token}).} 
Standard tokenization splits an input into patches using a regular grid, breaking shift-equivariance. We propose a data-dependent alternative that selects patches that maximize a shift-invariant function, resulting in the same tokens regardless of input shifts.

Given an input $\vx \in \sR^{N}$ and a patch length $L$, Our adaptive tokenization is defined as
\begin{align}
\label{eq:adaptive_token}
    \text{\texttt{A-token}}(\vx)=&\ \mX^{(m^{\star})}\mE \in \sR^{\frac{N}{L}\times D}\\
\label{eq:atoken_invariant_function}
    \text{with } m^{\star}=&\ \argmax_{m\in \{0,\dots,L-1\}} F(\mX^{(m)}\mE).
\end{align}
Here, $\mX^{(m)}=\text{\texttt{reshape}}(\gS_{N}^{m}\vx)\in \sR^{\frac{N}{L}\times L}$ is the reshaped version of the input circularly shifted by $m$ samples, $\mE \in \sR^{L\times D}$ is a linear projection and $F:\sR^{\frac{N}{L}\times D}\mapsto \sR$ is a shift-invariant function.
Note that the token representation of an input is only affected by circular shifts up to the patch size $L$. For any shift greater than $L-1$, there is a shift smaller than $L$ that generates the same tokens (up to a circular shift). So, an input has $L$ different token representations. \figref{fig:atoken} illustrates our proposed shift-equivariant tokenization.

Our adaptive tokenization maximizes a shift-invariant function to ensure the same token representation regardless of input shifts. Next, we analyze an essential property of $\mX^{(m)}\mE$ to prove that \texttt{A-token} is shift-equivariant.

\begin{mdframed}[style=MyFrame,align=center]
\begin{restatable}{lemma}{lem}
\label{clm:lemma}
\vspace{0.2em}
\underline{$L$-periodic shift-equivariance of tokenization}.

Let input $\vx\in \sR^{N}$ have a token representation $\mX^{(m)}\mE\in \sR^{\floor{N/L}\times D}$. If $\hat{\vx}=\gS_{N}\vx$ (a shifted input), then its token representation $\hat{\mX}^{(m)}\mE$ corresponds to:
\begin{align}
    \hat{\mX}^{(m)}\mE=&\ {\color{mybb}\gS_{\floor{N/L}}^{\floor{(m+1)/L}}} \mX^{(\circshift{(m+1)}{L})}\mE
\end{align}
This implies that $\vx$ and $\hat{\vx}$ are characterized by the same $L$ token representations, up to a {\color{mybb} circular shift} along the token index (row index of $\mX^{(\circshift{(m+1)}{L})}\mE$).
\end{restatable}
\end{mdframed}
\begin{proof}
\vspace{-0.2cm}
By definition, $\hat{\mX}^{(m)}=\text{\texttt{reshape}}(\gS_{N}^{m+1}\vx)$. Expressing $m+1$ in quotient and remainder for divisor $L$, we show that the remainder determines the relation between the $L$ token representations of an input and its shifted version, while the quotient causes a circular shift in the token index. The complete proof is deferred to Appendix~\secref{supp_sec:proof}.
\end{proof}

Lemma~\ref{clm:lemma} shows that, for any index $m$, there exists $\hat{m}= \circshift{(m+1)}{L}$ such that $\mX^{(m)}$ and $\hat{\mX}^{(\hat{m})}$ are equal up to a circular shift. In Claim \ref{clm:a_token_shifteq}, we use this property to demonstrate the shift-equivariance of our proposed adaptive tokenization.

\begin{mdframed}[style=MyFrame,align=center]
\begin{restatable}{claim}{token}
\label{clm:a_token_shifteq}
\underline{Shift-equivariance of adaptive tokenization}.

If $F$ in \equref{eq:atoken_invariant_function} is shift-invariant, then \texttt{A-token} is shift-equivariant, \ie, $\exists\ m_{q}\in \{0,\dots,L-1\}$ s.t.
\begin{align}
    \label{eq:a_token_shifteq}
    \text{\texttt{A-token}}\big(\gS_{N}\vx\big)=&\ \gS_{\floor{N/L}}^{m_{q}}\text{\texttt{A-token}}(\vx).
\end{align}
\end{restatable}
\end{mdframed}
\begin{proof}
\vspace{-0.2cm}

Given $m^{\star}$ in \equref{eq:atoken_invariant_function}, Lemma~\ref{clm:lemma} asserts the existence of $\hat{m}$ such that $\hat{\mX}^{(\hat{m})}\mE=\mX^{(m^{\star})}\mE$ up to a circular shift. Since $\vx$ and $\gS_{N}\vx$ have the same $L$ token representations and assuming a shift-invariant $F$, we show that \texttt{A-token}$(\gS_{N}\vx)$ is equal to $\hat{\mX}^{(\hat{m})}\mE$, which is a circularly shifted version of \texttt{A-token}$(\vx)$. See Appendix~\secref{supp_sec:proof} for the full proof.
\vspace{-0.1cm}
\end{proof}

{\noindent \bf Adaptive window-based self-attention (\texttt{A-WSA}).}
\texttt{WSA}'s window partitioning is shift-sensitive, as different windows are obtained when the input tokens are circularly shifted by a \textit{non-multiple of the window size}. Instead, we propose an adaptive token shifting method to obtain a consistent window partition. By selecting the offset based on the \textit{energy} of each possible partition, our method generates the same windows regardless of input shifts.

Given input tokens $\mT=\begin{bmatrix}\mT_{0} & \dots & \mT_{M-1}\end{bmatrix}^{\top}\in \sR^{M\times D}$ and a window size $W$, let $\vv_{W}\in \sR^{\floor{\frac{M}{W}}}$ denote the average $\ell_{p}$-norm (energy) of each token window:
\begin{align}
    \label{eq:awsa_energy_shift0}
    \vv_{W}[k]=&\ \frac{1}{W}\sum_{l=0}^{W-1}\|\mT_{\circshift{(Wk+l)}{M}}\|_{p}.
\end{align}
Then, the energy of the windows resulting from shifting the input tokens by $m$ indices corresponds to $\vv_{W}^{(m)}\in \sR^{\floor{\frac{M}{W}}}$, where $\vv_{W}^{(m)}[k]$ is the energy of the $k^{\text{th}}$ window:
\begin{align}
    \label{eq:awsa_energy_shift}
    \vv_{W}^{(m)}[k]=& \frac{1}{W}\sum_{l=0}^{W-1}\|\underset{=\mT_{\circshift{(Wk+m+l)}{M}}}{\underbrace{(\gS_{M}^{m}\mT)_{\circshift{(Wk+l)}{M}}}}\|_{p}.
\end{align}

Based on the window energy in~\equref{eq:awsa_energy_shift}, we define the adaptive window-based self-attention as
\begin{align}
    \label{eq:a_wsa}
    \text{\texttt{A-WSA}}(\mT)=\ \text{\texttt{WSA}}\big(\gS_{M}^{m^{\star}}\mT\big)\in \sR^{M \times D'}\\
    \text{with } m^{\star}= \argmax_{m\in \{0,\dots,W-1\}}G\big(\vv_{W}^{(m)}\big)
\end{align}
where $G:\sR^{\floor{\frac{M}{W}}}\mapsto \sR$ is a shift-invariant function. By choosing windows based on $m^{\star}$, \texttt{A-WSA} generates the same group of windows despite input shifts, as shown in Claim~\ref{claim:awsa_equivariance}.

\begin{mdframed}[style=MyFrame,align=center]
\begin{restatable}{claim}{awsa}
\label{claim:awsa_equivariance}
If $G$ in~\equref{eq:a_wsa} is shift invariant, then \texttt{A-WSA} is shift-equivariant.
\end{restatable}
\end{mdframed}
\begin{proof}
\vspace{-0.1cm}
Given two groups of tokens related by a circular shift, and a shift-invariant function $G$, shifting each group by its maximizer in~\equref{eq:a_wsa} induces an offset that is a multiple of $W$. So, both groups are partitioned in the same windows up to a circular shift. \figref{fig:awsa} illustrates this consistent window grouping. The proof is deferred to Appendix~\secref{supp_sec:proof}.
\end{proof}

\begin{figure*}[h]
    \centering
    \subfloat[Shifted tokens]{
    \includegraphics[height=0.15\linewidth,valign=t]{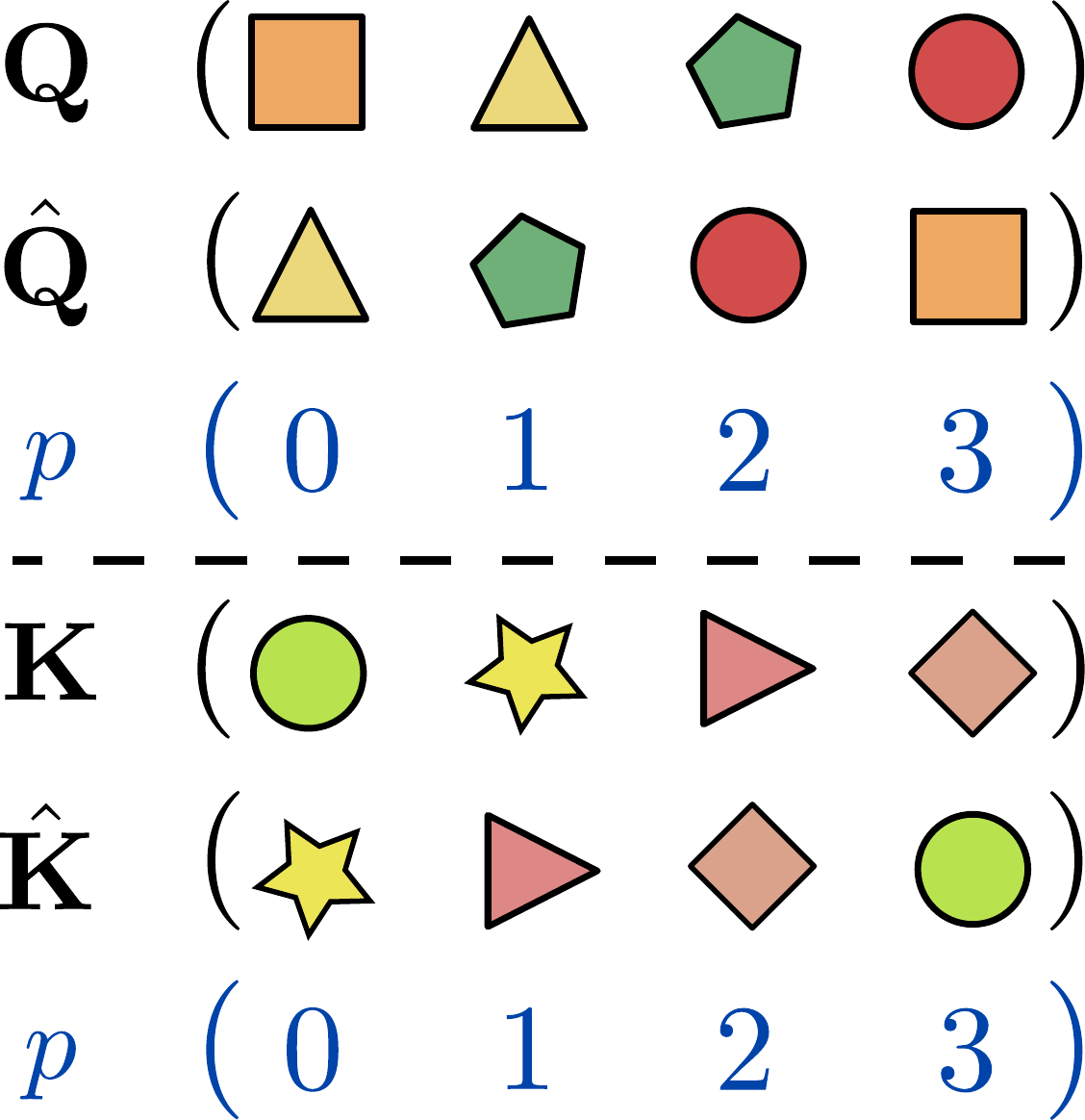}\label{fig:rpe_input}\hspace{0.05\linewidth}}
    \subfloat[Original relative distance]{\includegraphics[height=0.16\linewidth,valign=t]{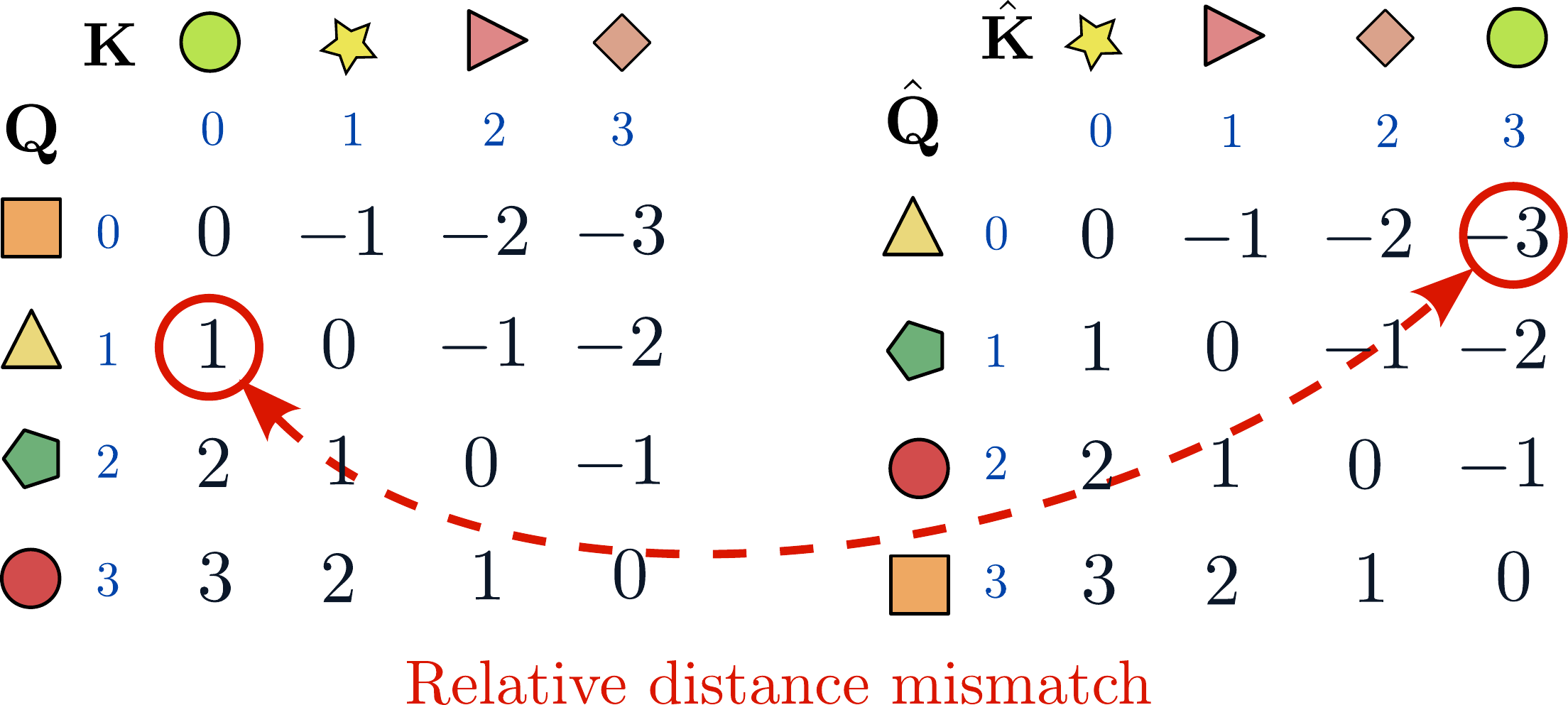}\label{fig:rpe_orig}\hspace{0.09\linewidth}}
    \subfloat[Proposed relative distance]{\includegraphics[height=0.16\linewidth,valign=t]{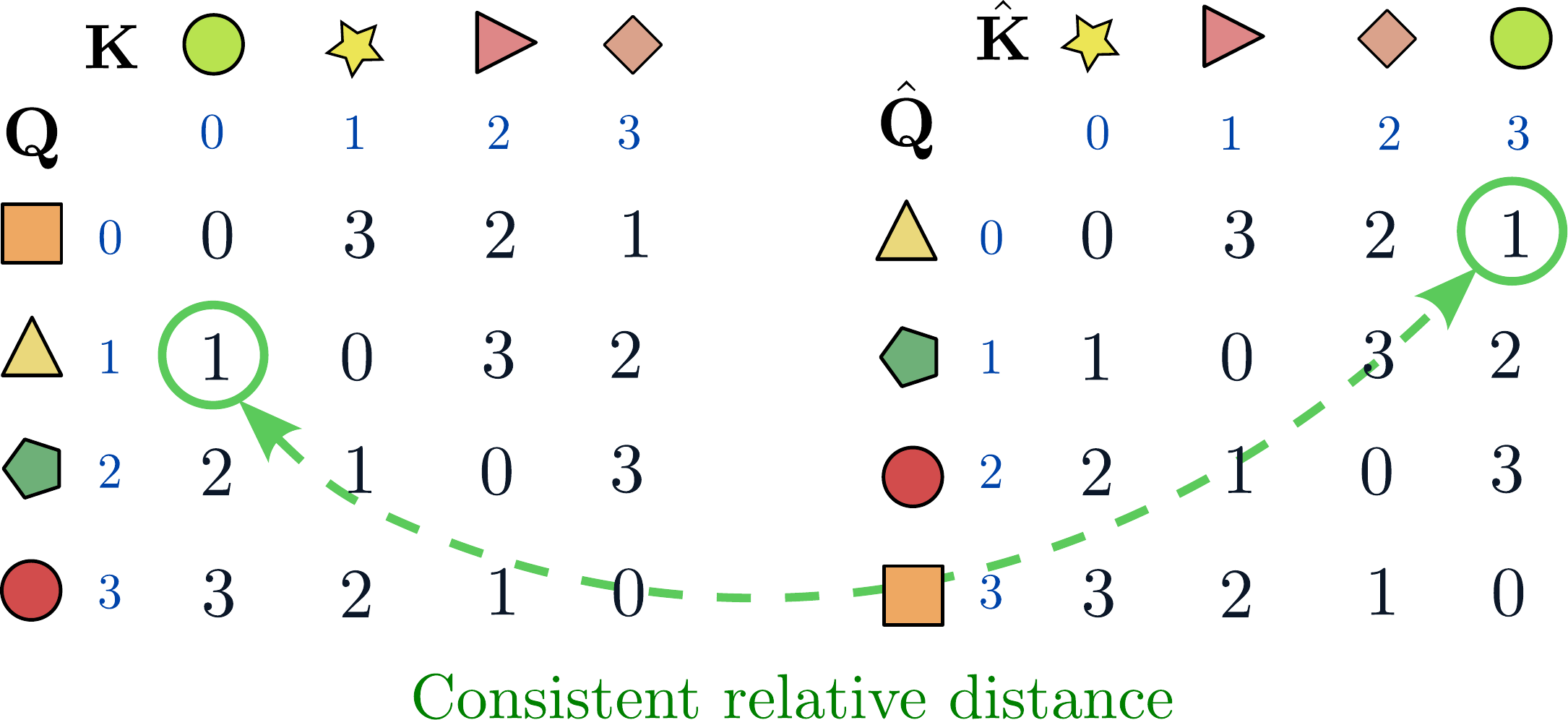}\label{fig:arpe}}
    \vspace{-0.2cm}
    \caption{\textbf{Shift consistent relative distance:} (a) Circularly shifted queries and keys $(M=4)$. (b) Original relative distance used to build the RPE matrix: $p^{(Q)}[i]-p^{(K)}[j]$. Since it does not consider the periodicity of circular shifts, relative distances are not preserved. (c) Proposed relative distance: $(p^{(Q)}[i]-p^{(K)}[j]) \text{mod}\ M$. Our proposed distance is consistent with circular shifts, leading to a shift equivariant RPE.}
    \label{fig:rpe}
\end{figure*}

\myparagraph{\noindent \bf Adaptive patch merging (\texttt{A-PMerge}).}
As reviewed in~\secref{sec:prelim}, \texttt{PMerge} consists of a vectorization of $P$ neighboring tokens followed by a projection from $\sR^{PD}$ to $\sR^{\tilde{D}}$. So, it can be expressed as a strided convolution with $\tilde{D}$ output channels, stride factor $P$ and kernel size $P$. We use this property to propose a shift-equivariant patch merging.
\begin{mdframed}[style=MyFrame,align=center]
\begin{restatable}{claim}{pmergeconv}
\label{claim:pmergeconv}
\texttt{PMerge} corresponds to a strided convolution with $\tilde{D}$ output channels, striding $P$ and kernel size $P$.
\end{restatable}
\end{mdframed}

\begin{proof}
\vspace{-0.3cm}
Expressing the linear projection $\tilde{\mE}$ as a convolutional matrix, \texttt{PMerge} is equivalent to a convolution sum with kernels comprised by columns of $\tilde{\mE}$.
Let the input tokens be expressed as $\mT= \begin{bmatrix}\vt_{0} & \dots & \vt_{D-1}\end{bmatrix} \in \sR^{M\times D}$, where $\vt_{j}\in \sR^{M}$ corresponds to the $j^{\text{th}}$ element of every input token. Then, $\text{\texttt{PMerge}}(\mT)\in \sR^{\frac{M}{P}\times \tilde{D}}$ can be expressed as:
\begin{align}
    \text{\texttt{PMerge}}(\mT)=&\ \gD^{(P)}(\begin{bmatrix}\vy_{0} & \dots & \vy_{\tilde{D}-1}\end{bmatrix})\\
    \text{with } \vy_{k} =&\ \sum_{j=0}^{D-1}\vt_{j}\circledast \vh^{(k,j)} \in \sR^{M}
\end{align}
where $\gD^{(P)}\in \sR^{\frac{M}{P}\times M}$ is a striding operator of factor $P$, $\circledast$ denotes circular convolution and $\vh^{(k,j)}\in \sR^{P}$ is a kernel. Details are deferred to Appendix~\secref{supp_sec:proof}.
\vspace{-0.2cm}
\end{proof}

Following Claim~\ref{claim:pmergeconv}, to attain shift-equivariance, the adaptive polyphase sampling (APS) by~\citet{chaman2021truly} is adopted as the striding operator. Let \texttt{APS}$^{(P)}$ denote the adaptive polyphase sampling layer of striding factor $P$. Then, $\text{\texttt{A-PMerge}}(\mT)\in \sR^{\frac{M}{P}\times \tilde{D}}$ corresponds to:
\begin{align}
    \text{\texttt{A-PMerge}}(\mT)=&\ \text{\texttt{APS}}^{(P)}(\begin{bmatrix}\vy_{0} & \dots & \vy_{\tilde{D}-1}\end{bmatrix})
\end{align}
While \texttt{PMerge} selects $\frac{M}{P}$ embeddings by striding, shift-equivariance is achieved by adaptively choosing the best $\frac{M}{P}$ tokens based on their $\ell_{2}$ norm.

\myparagraph{\noindent \bf Adaptive RPE.}
While the original relative distance matrix $\mE^{(\text{rel})}$ is computed by taking into account linear shifts, this does not match our circular shift assumption; See~\figref{fig:rpe} for a visualization. To obtain perfect shift-equivariance, relative distances must consider the periodicity induced by circular shifts. Hence, we propose the adaptive relative position matrix $\mE^{(\text{adapt})}\in \sR^{M\times M}$ defined as:
\begin{align}
    \mE^{(\text{adapt})}[i,j]=&\ \mB^{(\text{adapt})}\left[\circshift{(p^{(Q)}_{i}-p^{(K)}_{j})}{M}\right]
\end{align}
to encode the distance between the $i^{\text{th}}$~query token at position $p^{(Q)}_{i}$ and the $j^{\text{th}}$~key token at position $p^{(K)}_{j}$. Here, $\mB^{\text{(adapt)}}\in \sR^{M}$ is the trainable lookup table comprised by relative positional embeddings. Note that $\mB^{(\text{adapt})}$ is smaller than the original $\mB^{(\text{rel})}\in \sR^{(2M-1)}$, since relative distances are now measured in a circular fashion between $M$ tokens.

\myparagraph{Segmentation with equivariant upsampling.}
Segmentation models with ViT backbones continue to use CNN decoders, \eg, Swin~\cite{liu_2021_swin} uses UperNet~\cite{xiao2018unified}. As explained by~\citet{rojas-neurips2022-learnable}, to obtain a shift-equivariant CNN decoder, the key is to keep track of the downsampling indices and use them to put features back to their original positions during upsampling. Different from CNNs, the proposed ViTs involve data-adaptive window selections,~\ie, we would also need to keep track of the window indices and account for their shifts during upsampling.

\section{Experiments}
\label{sec:experiments}
We conduct experiments on image classification and semantic segmentation on four ViT architectures, namely, Swin~\cite{liu_2021_swin}, SwinV2~\cite{liu2022swin}, CvT~\cite{wu2021cvt}, and MViTv2~\cite{li2022mvitv2}. For each task, we analyze their performance under circular and standard shifts. For circular shifts, the theory matches the experiments and our approach achieves 100\% circular shift consistency (up to numerical errors). We further conduct experiments on standard shifts to study how our method performs under this theory-to-practice gap, where there is loss of information at the boundaries.

\begin{table*}[t]
\centering
\small
\setlength{\tabcolsep}{4.5pt}

\rowcolors{1}{bg_blue}{}
\resizebox{\textwidth}{!}{%
\begin{tabular}{ccc cc|cc cc}
\specialrule{.15em}{.05em}{.05em}
\hiderowcolors
\multirow{3}{*}{Method} & \multicolumn{4}{c|}{Circular Shift} & \multicolumn{4}{c}{Standard Shift}\\
 & \multicolumn{2}{c}{CIFAR10} & \multicolumn{2}{c|}{CIFAR-100} & \multicolumn{2}{c}{CIFAR10} & \multicolumn{2}{c}{CIFAR-100}\\
 & Top-1 Acc.   & C-Cons.   & Top-1 Acc.   & C-Cons.   & Top-1 Acc.   & S-Cons. & Top-1 Acc.   & S-Cons. \\
 \midrule
 \showrowcolors 
   Swin-T     & $90.15\pm.18$ & $83.30\pm.61$             
           & $71.01\pm.27$ & $65.32\pm.69$ & $90.11\pm.21$ & $86.35\pm.25$
           & $71.12\pm.14$ & $69.39\pm.52$ \\
  A-Swin-T \bf(Ours)  & $\bf 93.39\pm.12$ & $\bf 99.99\pm.01 $    
           & $\bf 75.11\pm.10$ & $\bf 99.99\pm.01$ & $\bf 93.50\pm.19$ & $\bf 96.00\pm.08$
           & $\bf 75.12\pm.28$ & $\bf 87.70\pm.57$\\
  SwinV2-T   & $89.08\pm.21$ & $89.16\pm.08$          
           & $69.78\pm.22$ & $75.23\pm.20$ & $89.08\pm.21$ & $91.68\pm.25$
           & $69.67\pm.32$ & $80.42\pm.41$  \\
  A-SwinV2-T \bf (Ours) & $\bf 91.64\pm.21$ & $\bf 99.99\pm.01$  
           & $\bf 72.73\pm.23$ & $\bf 99.96\pm.01$  & $\bf 91.91\pm.12$ & $\bf 95.81\pm.17$
           & $\bf 72.98\pm.13$ & $\bf 88.74\pm.40$ \\
  CvT-13      & $90.06\pm.23$ & $75.80\pm1.2$            
           & $66.61\pm.33$ & $50.29\pm1.68$ & $90.05\pm.20$ & $84.66\pm1.26$
           & $66.06\pm.39$ & $63.03\pm.73$ \\
  A-CvT-13 \bf (Ours)    & $\bf 93.87 \pm .14$ & $\bf 100 \pm .00$     
           & $\bf 76.19 \pm.32$ & $\bf 100 \pm .00$ & $\bf 93.71\pm.10$ & $\bf 96.47\pm.21$
           & $\bf 73.04\pm.23$ & $\bf 86.96.\pm.55$\\
  MViTv2-T     & $96.00\pm.06$ & $86.55\pm1.2$            
           & $80.18\pm.34$ & $74.82\pm.73$ & $96.14\pm.06$ & $91.34.\pm1.26$
           & $80.28\pm.38$ & $77.92.\pm.93$ \\
  A-MViTv2-T \bf (Ours)   & $\bf 96.41 \pm .22$ & $\bf 100 \pm .00$     
           & $\bf 81.39\pm.11$ & $\bf 100 \pm .00$ & $\bf 96.61\pm.11$ & $\bf 98.36.\pm.16$
           & $\bf 81.17\pm.18$ & $\bf 92.95.\pm.16$\\
\specialrule{.15em}{.05em}{.05em}
\end{tabular}
}
\vspace{-0.2cm}
\caption{\textbf{CIFAR-10/100 classification results:} Top-1 accuracy and shift consistency under circular and linear shifts. Bold numbers indicate improvement over the corresponding baseline architecture. Mean and standard deviation reported on five randomly initialized models.
}
\label{tab:class_cifar}
\end{table*}

\subsection{Image classification under circular shifts}
{\bf \noindent Experiment setup.}
We conduct experiments on CIFAR-10/100~\cite{krizhevsky2009learning}, and ImageNet~\cite{deng2009imagenet}. For all datasets, images are resized to the resolution used by each model's original implementation ($224 \times 224$ for Swin-T, CVT-13, and MViTv2-T; $256 \times 256$ for SwinV2-T). Using the default image size allows us to use the same architecture across all datasets, \ie, everything follows the original number of layers and blocks. To avoid boundary conditions, circular padding is used in all convolutional layers, and circular shifts are used for evaluating shift consistency.

On CIFAR-10/100, all models were trained for $100$ epochs on two GPUs with batch size $48$. The scheduler settings of each model were scaled accordingly. On ImageNet, all models were trained for $300$ epochs on eight GPUs using their default batch sizes. Refer to~\secref{sec:supp_exp_details} for full experimental details. For CIFAR-10/100, we report average and standard deviation metrics over five seeds. Due to computational limitations, we report on a single seed for ImageNet.

\myparagraph{Evaluation metric.}
We report the classification top-1 accuracy on the original dataset without any shifts. To quantify shift-invariance, we also report the circular shift consistency (C-Cons.) which counts how often the predicted labels are identical under two different circular shifts. Given a dataset $\gD=\{\mI\}$, C-Cons. computes
\begin{align}
\label{eq:ccons}
\frac{1}{|\gD|}\sum_{\mI \in \gD} \mathbb{E}_{\bm{\Delta}_{1},\bm{\Delta}_{2}}  {\color{mybb}\Big[} \mathbf{1} {\color{myrr}\big[}\hat{y}(\gS ^{\bm{\Delta}_{1}}(\mI)) = \hat{y}(\gS^{\bm{\Delta}_{2}}(\mI)){\color{myrr}\big]} {\color{mybb}\Big]}
\end{align}
where $\mathbf{1}$ denotes the indicator function, $\hat{y}(\mI)$ the class prediction for $\mI$, $\gS$ the circular shift operator, and $\bm{\Delta}_{1}=(h_1, w_1), \bm{\Delta}_{2}=(h_2, w_2)$ horizontal and vertical offsets.

{\bf \noindent Results.} We report performance in~\tabref{tab:class_cifar} and~\tabref{tab:class_imagenet} for CIFAR-10/100 and ImageNet, respectively. Overall, we observe that our adaptive ViTs achieve near 100\% shift consistency in practice. The remaining inconsistency is caused by numerical precision limitations and tie-breaking leading to a different selection of tokens or shifted windows. Beyond consistency improvements, our method also improves classification accuracy across all settings.

\begin{table}[t]
\centering
\small
\setlength{\tabcolsep}{2pt}

\rowcolors{1}{}{bg_blue}
\resizebox{\columnwidth}{!}{%
\begin{tabular}{ccc|cc}
\specialrule{.15em}{.05em}{.05em}
\hiderowcolors 
\multirow{2}{*}{Method} & \multicolumn{2}{c}{Circular Shift} & \multicolumn{2}{c}{Standard Shift}\\
 & Top-1 Acc.   & C-Cons.   & Top-1 Acc.   & S-Cons.\\
 \midrule
 \showrowcolors 
   Swin-T     & $78.5$ & $86.68$ & $81.18$ & $92.41$\\
  A-Swin-T \bf(Ours)  & $\bf 79.35$ & $\bf 99.98$ & $\bf 81.6$ & $\bf 93.24$\\
  SwinV2-T   & $78.95$ & $87.68$ & $81.76$ & $93.24$\\
  A-SwinV2-T \bf (Ours) & $\bf 79.91$ & $\bf 99.98$ & $\bf 82.10$ & $\bf 94.04$\\
  CvT-13      & $77.01$ & $86.87$ & $\bf 81.59$ & $92.80$\\
  A-CvT-13 \bf (Ours)    & $\bf 77.05$ & $\bf 100$ & $81.48$ & $\bf 93.41$\\
  MViTv2-T     & $77.36$ & $90.03$ & $82.21$ & $93.88$\\
  A-MViTv2-T \bf (Ours)   & $\bf 77.46$ & $\bf 100$ & $\bf 82.4$ & $\bf 94.08$\\
\specialrule{.15em}{.05em}{.05em}
\end{tabular}
}
\vspace{-0.2cm}
\caption{\textbf{ImageNet classification results:} Top-1 accuracy and shift consistency under circular and linear shifts. Bold numbers indicate improvement over the corresponding baseline architecture.
}
\vspace{-0.6cm}
\label{tab:class_imagenet}
\end{table}
\subsection{Image classification under standard shifts}
\label{sec:classification}
{\bf \noindent Experiment setup.} To study the boundary effect on shift-invariance, we further conduct experiments using standard shifts. As these are no longer circular, the image content may change at its borders, \ie, perfect shift consistency is \textit{no longer guaranteed}. For CIFAR10/100, input images were resized to the resolution used by each model's original implementation. Default data augmentation and optimizer settings were used for each model while training epochs and batch size followed those used in the circular shift settings.

\myparagraph{Evaluation metric.}
We report top-1 classification accuracy on the original dataset (without any shifts). To quantify shift-invariance, we report the standard shift consistency (S-Cons.), which follows the same principle as C-Cons in~\equref{eq:ccons}, but uses a standard shift instead of a circular one. For CIFAR-10/100, we use zero-padding at the boundaries. due to the small image size. For ImageNet, following~\citet{zhang2019making}, we perform an image shift followed by a center-cropping of size $224 \times 224$. This produces realistic shifts and avoids a particular choice of padding.

\begin{figure*}[h]
\colorbox{white}{\begin{minipage}[t]{0.095\textwidth}
\centering \textbf{\scalebox{0.6}{\makecell{Input\\ $(224 \times 224)$}}}
\end{minipage}}\colorbox{white}{\begin{minipage}[t]{0.14\textwidth}
\centering \textbf{\scalebox{0.6}{\makecell{CvT-13\\ Block $\bm 1$ Error $(56 \times 56)$}}}
\end{minipage}}\colorbox{white}{\begin{minipage}[t]{0.13\textwidth}
\centering \textbf{\scalebox{0.6}{\textcolor{black}{\makecell{A-CvT-13 (Ours)\\ Block $1$ Error $(56 \times 56)$}}}}
\end{minipage}}\colorbox{white}{\begin{minipage}[t]{0.135\textwidth}
\centering \textbf{\scalebox{0.6}{\makecell{CvT-13\\ Block $\bm 2$ Error $(28 \times 28)$}}}
\end{minipage}}\colorbox{white}{\begin{minipage}[t]{0.14\textwidth}
\centering \textbf{\scalebox{0.6}{\textcolor{black}{\makecell{A-CvT-13 (Ours)\\ Block $2$ Error $(28 \times 28)$}}}}
\end{minipage}}\colorbox{white}{\begin{minipage}[t]{0.145\textwidth}
\centering \textbf{\scalebox{0.6}{\makecell{CvT-13\\ Block $\bm 3$ Error $(14 \times 14)$}}}
\end{minipage}}\colorbox{white}{\begin{minipage}[t]{0.12\textwidth}
\centering \textbf{\scalebox{0.6}{\textcolor{black}{\makecell{A-CvT-13 (Ours)\\ Block $3$ Error $(14 \times 14)$}}}}
\end{minipage}}

\vspace{-0.2 cm}

\centering
\subfloat{\includegraphics[width=1\textwidth]{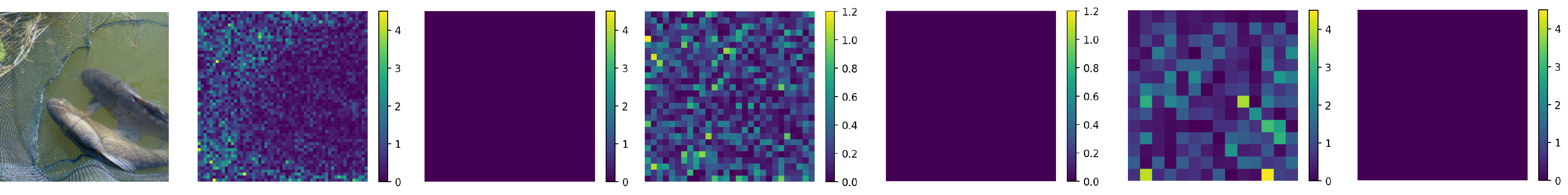}}
\vspace{-0.35cm}
\caption{\textbf{Consistent token representations.} %
Shifting inputs by a small offset leads to large deviations (non-zero errors) in the representations when using default ViTs (\eg, CvT-13). In contrast, our proposed models (\eg, A-CvT-13) achieve an absolute zero-error across all blocks.
}
\label{fig:rep_stability}
\vspace{-0.3cm}
\end{figure*}

\myparagraph{Results.}
Tabs. \ref{tab:class_cifar} and \ref{tab:class_imagenet} report performance under standard shifts on CIFAR-10/100 and ImageNet, respectively. Due to changes in the boundary content, our method does not achieve 100\% shift consistency. However, we persistently observe that the adaptive models outperform their respective baselines in terms of S-Cons. Beyond shift consistency, our adaptive models achieve higher classification performance in all settings except for CvT on ImageNet. Results demonstrate the practical value of our approach despite the gap in theory.

\subsection{Consistency of tokens to input shifts}
We evaluate the effect of small input shifts in the tokens obtained by our adaptive models. We verify the stability of our A-CvT-13 model by applying a circular shift of $1$ row and $1$ column to the input image, computing its tokens, and calculating their absolute difference to those of the unshifted image. \figref{fig:rep_stability} shows the absolute token difference of an ImageNet test sample at all three blocks of A-CvT-13, each with a different resolution. Similar to previous work \cite{chaman2021truly}, we illustrate errors for the channels with the highest energy.

In contrast to the large deviations across the default CvT-13 caused by the input shift, the representations generated by our proposed A-CvT-13 model remain unaltered, as theoretically shown, leading to a perfectly shift-equivariant model.

\begin{table}[t]
\centering
\small
\setlength{\tabcolsep}{3.5pt}
\rowcolors{1}{bg_blue}{}
\begin{tabular}{cccc}
\toprule
\hiderowcolors 
Model & \makecell{\# Params} & \makecell{Throughput\\ (images/s)} & \makecell{Relative\\ change $(\%)$}\\ 
\midrule
\showrowcolors
Swin-T & $28$M & $704.07$ & $-$\\
A-Swin-T \textbf{(Ours)} & $28$M & $633.35$& $10.04$\\
SwinV2-T & $28$M & $470.81$ & $-$ \\
A-SwinV2-T  \textbf{(Ours)} & $28$M & $405.01$ & $13.98$\\
CvT-13 & $20$M & $535.5$ & $-$\\
A-CvT-13 \textbf{(Ours)} & $20$M & $492.12$ & $10.69$ \\
MViTv2-T & $24$M & $439.5$ & $-$ \\
A-MViTv2-T \textbf{(Ours)} & $24$M & $352.06$ & $19.9$\\
\bottomrule
\end{tabular}
\vspace{-0.2cm}
\caption{\textbf{Inference throughput:} Absolute inference throughput (images/s) of our adaptive ViTs and their default versions. %
\emph{Relative change} shows the throughput decrease w.r.t. the default models.}
\label{tab:throughput_inference}
\vspace{-0.2cm}
\end{table}

\begin{table}
\centering
\small
\setlength{\tabcolsep}{0.5pt}

\rowcolors{1}{bg_blue}{}
\resizebox{\columnwidth}{!}{%
\begin{tabular}{cccc}
\toprule
\hiderowcolors 
Module & Abs. runtime (ms) & Delta (ms)\\ 
\midrule
\showrowcolors
Tokenization & $8.37$ & $-$\\
A. Tokenization \textbf{(Ours)} & $35.89$ & $+27.52$ \\
Patch Merging $\{$S2, S3, S4$\}$ & $0.47$, $0.45$, $0.45$ & $-$ \\
A. Patch Merging \textbf{(Ours)} & $7.68$, $4.47$, $3.09$ & $+7.21$, $+4.02$, $+2.64$ \\
Window Selection & Not applied & $-$ \\
A. Window Selection \textbf{(Ours)} & $7.63$ & $+7.63$\\
RPE & $2.84$ & $-$ \\
A. RPE \textbf{(Ours)} & $9.91$ & $+7.07$ \\
\bottomrule
\end{tabular}
}
\vspace{-0.2cm}
\caption{\textbf{Runtime of adaptive ViT modules:} Inference runtime of our adaptive ViT modules and their default versions. %
\textit{Delta} indicates the absolute time difference w.r.t. the default modules.}
\label{tab:runtime_modules}
\vspace{-0.3 cm}
\end{table}

\subsection{Throughput and runtime analysis}
We evaluate the inference throughput, measured in processed images per second, of our adaptive ViTs and modules over $100$ forward passes (batch size $128$, default image size per model) on a single NVIDIA Quadro RTX 5000 GPU.

\myparagraph{Model inference.} We report the throughput of our adaptive ViTs and their default versions. We also measure the relative change, which corresponds to the throughput decrease with respect to the default models.

\tabref{tab:throughput_inference} shows our adaptive models exhibit less than a 20\% decrease in throughput w.r.t. the default models, while improving in shift consistency and classification accuracy without increasing the number of trainable parameters.

\myparagraph{Modules runtime.}
We compare the runtime of our adaptive modules to that of the default ones. Tokenization, patch merging and window selection are evaluated on A-Swin, while RPE is evaluated on A-MViTv2 (A-Swin windows are comprised of the same tokens, so its RPE remains unaltered).

\tabref{tab:runtime_modules} shows the runtime of our adaptive modules, which slightly increases over the default runtime. This is particularly true for the adaptive RPE, where the main difference lies in the distance interpretation (circular vs. linear). While adaptive patch embedding has the largest increase by operating on full-size images, subsequent patch merging modules operate on smaller representations and are more efficient.

\subsection{Semantic segmentation under circular shifts}
{\bf \noindent Experiment setup.}
We conduct semantic segmentation experiments on the ADE20K dataset~\cite{zhou2017scene} using A-Swin and A-SwinV2 models as backbones and compare them against their default versions. Following previous work \citep{liu_2021_swin}, we use UperNet~\cite{xiao2018unified} as the segmentation decoder. Similar to our classification settings under circular shifts, all convolutional layers in the UperNet model use circular padding to avoid boundary conditions, and circular shifts are used to measure shift consistency. Models are trained for $160$K iterations on a total batch size of $16$ using the default augmentation.

\begin{figure*}[t]
\colorbox{white}{\begin{minipage}[t]{0.325\columnwidth}
\centering \textbf{\scalebox{0.8}{Image}}
\end{minipage}}\colorbox{white}{\begin{minipage}[t]{0.325\columnwidth}
\centering \textbf{\scalebox{0.8}{Shifted Image}}
\end{minipage}}\colorbox{white}{\begin{minipage}[t]{0.67\columnwidth}
\centering \textbf{\scalebox{0.8}{SwinV2 $+$ UperNet Predictions}}
\end{minipage}}\colorbox{white}{\begin{minipage}[t]{0.67\columnwidth}
\centering\textbf{\scalebox{0.8}{A-SwinV2 $+$ UperNet Predictions (Ours)}}
\end{minipage}}

\vspace{-0.1 cm}
\includegraphics[width=1\textwidth]{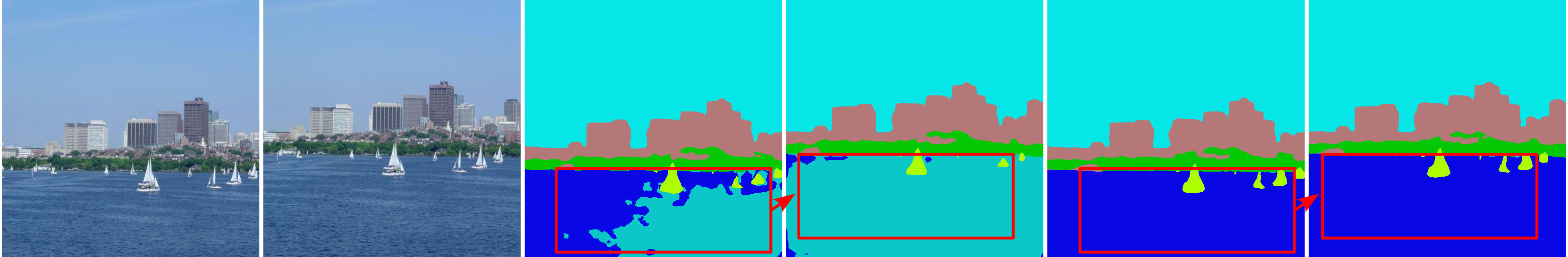}

\vspace{-0.17 cm}
\caption{\label{fig:seg_qual}\textbf{Segmentation under standard shifts:} Our A-SwinV2 $+$ UperNet model improves robustness to input shifts over the original model, generating consistent predictions while improving accuracy. Examples of prediction changes due to shifts are highlighted in red.}
\vspace{-0.15cm}
\end{figure*}

\begin{table}
\centering
\small
\setlength{\tabcolsep}{3.5pt}
\rowcolors{2}{}{bg_blue}
\begin{tabular}{ccc|cc}
\toprule
\hiderowcolors 
\multirow{2}{*}{Backbone}& \multicolumn{2}{c|}{Circular Shift} & \multicolumn{2}{c}{Standard Shift}\\
& mIoU   & mASCC   & mIoU   & mASSC  \\ 
\midrule
\showrowcolors
Swin-T & $42.93$ & $87.32$ & $44.2$ & $93.37$\\
A-Swin-T \textbf{(Ours)} & $\bf 43.44$ & $\bf 100$ & $\bf44.43$ & $\bf 93.48$\\
SwinV2-T & $43.86$ & $88.16$ & $44.26$ & $93.23$\\
A-SwinV2-T \textbf{(Ours)} & $\bf 44.42$ & $\bf 100$ & $\bf 46.11$ & $\bf 93.59$\vspace{-0.1 cm}\\
\bottomrule
\end{tabular}
\vspace{-0.2cm}
\caption{\textbf{Semantic segmentation performance:} Segmentation accuracy and shift consistency of our adaptive UperNet model equipped with A-Swin and A-SwinV2 backbones.}
\label{tab:seg_acc_combined}
\vspace{-0.2cm}
\end{table}

\myparagraph{Evaluation metric.}
For segmentation performance, we report the mean intersection over union (mIoU) on the original dataset (without any shifts). For shift-equivariance, we report the mean-Average Segmentation Circular Consistency (mASCC) which counts how often the predicted pixel labels~(after shifting back) are identical under two different circular shifts. Given a dataset $\gD=\{\mI\}$, mASCC computes
\begin{align}
\label{eq:mascc}
\small
\nonumber
&
\hspace*{-0.3cm}
\frac{1}{|\gD|}\sum_{\mI \in \gD} \mathbb{E}_{\bm{\Delta}_1,\bm{\Delta}_2}{\color{mybb}\Bigg[} 
\frac{1}{HW}\sum_{u=1,v=1}^{H,W} \mathbf{1} 
{\color{myrr}\Big[}\\
&\gS^{-\bm{\Delta}_{1}}\hat{y}(\gS^{\bm{\Delta}_{1}}(\mI))[u,v]= \gS^{-\bm{\Delta}_{2}}\hat{y}(\gS^{\bm{\Delta}_{2}}(\mI))[u,v]{\color{myrr}\Big]} {\color{mybb}\Bigg]},
\end{align}
where $H,W$ correspond to the image height and width, and $[u,v]$ indexes the class prediction at pixel $(u,v)$.

\myparagraph{Results.} \tabref{tab:seg_acc_combined} shows classification accuracy and shift consistency for UperNet segmenters using Swin-T and SwinV2-T backbones. Following the theory, our adaptive models achieve 100\% mASCC (perfect shift consistency), while improving on segmentation accuracy. 

\subsection{Semantic segmentation under standard shifts}
{\bf \noindent Experiment setup.} As in the circular shift scenario, models are trained for $160$K iterations with a total batch size of $16$ using the default data augmentation. To evaluate shift-equivariance under standard shifts, we report the mean-Average Semantic Segmentation Consistency (mASSC), which computes whether the percentage of predicted pixel labels~(after shifting back) remains under two standard shifts. Note, mASSC ignores the boundary pixels in its computation as standard shifts lead to changes in boundary content.

\myparagraph{Results.} \tabref{tab:seg_acc_combined} reports results on the standard shift scenario. Due to boundary conditions, perfect shift-equivariance is not achieved. Nevertheless, our models improve segmentation performance and shift consistency, with a notable improvement on SwinV2-T. See \figref{fig:seg_qual} for segmentation results.

\subsection{Ablation study}
We study the impact of each proposed ViT module by removing them. These include (i) adaptive tokenization, (ii) adaptive window-based self-attention, and (iii) adaptive patch merging. We conduct the ablations on our A-Swin-T model trained on CIFAR-10 under circular shifts.

Results are reported in \tabref{tab:ablation}. Our full model improves shift consistency by more than 3.5$\%$ over A-Swin-T without \texttt{A-token}, while slightly decreasing classification accuracy by 0.27$\%$. The use of \texttt{A-WSA} improves both classification accuracy and shift consistency. Finally, \texttt{A-PMerge} improves classification accuracy by approximately $1.7\%$ and shift consistency by more than $5\%$. Overall, all the proposed modules are necessary to achieve 100\% shift consistency.

\begin{table}
\centering
\small
\centering

\renewcommand{\arraystretch}{0.905}
\rowcolors{1}{}{bg_blue}
\begin{tabular}{ccc}
\toprule
\hiderowcolors 
Configuration & Top-1 Acc. & C-Cons.\\ 
\midrule
\showrowcolors
  A-Swin-T \textbf{(Ours)} & $93.39 \pm .13$ & $\bf 100$\\
\hiderowcolors
  (i) No \texttt{A-token} & $\bf 93.66 \pm .19$ & $96.29 \pm .20$\\
  (ii) No \texttt{A-WSA} & $93.24 \pm .15$ & $95.62 \pm .54$\\
  (iii) No \texttt{A-PMerge} & $91.67 \pm .10$ & $94.62 \pm .11$\\
\midrule
  Swin-T (Default) & $90.15 \pm .18$ & $83.30 \pm .61$\\
\bottomrule
\end{tabular}
\vspace{-0.2cm}
\caption{\textbf{Ablation study:} Effect of our shift-equivariant ViT modules on classification accuracy and shift consistency. Configurations progressively evaluated on Swin-T under circular shifts.}
\label{tab:ablation}
\vspace{-0.25 cm}
\end{table}

\section{Conclusion}
We propose a family of ViTs that are truly shift-invariant and equivariant. We redesigned four ViT modules, including tokenization, self-attention, patch merging, and positional embedding to guarantee circular shift-invariance and equivariance theoretically. Using these modules, we made Swin, SwinV2, CvT, and MViTv2 versions that are truly shift-equivariant. When matching the theoretical setup, these models exhibit 100\% circular shift consistency and superior performance compared to baselines. Furthermore, on standard shifts where image boundaries deviate from the theory, our proposed models remain more resilient to shifts with task performance on par with/exceeding the baselines.

\clearpage
\bibliographystyle{abbrvnat}
\bibliography{refs}

\begin{thebibliography}{57}
\providecommand{\natexlab}[1]{#1}
\providecommand{\url}[1]{\texttt{#1}}
\expandafter\ifx\csname urlstyle\endcsname\relax
  \providecommand{\doi}[1]{doi: #1}\else
  \providecommand{\doi}{doi: \begingroup \urlstyle{rm}\Url}\fi

\bibitem[Azulay and Weiss(2019)]{Azulay_Weiss_2019}
A.~Azulay and Y.~Weiss.
\newblock Why do deep convolutional networks generalize so poorly to small
  image transformations?
\newblock \emph{JMLR}, 2019.

\bibitem[Bronstein et~al.(2017)Bronstein, Bruna, LeCun, Szlam, and
  Vandergheynst]{bronstein2017geometric}
M.~M. Bronstein, J.~Bruna, Y.~LeCun, A.~Szlam, and P.~Vandergheynst.
\newblock Geometric deep learning: going beyond euclidean data.
\newblock \emph{IEEE SPM}, 2017.

\bibitem[Chaman and Dokmanic(2021)]{chaman2021truly}
A.~Chaman and I.~Dokmanic.
\newblock Truly shift-invariant convolutional neural networks.
\newblock In \emph{Proc. CVPR}, 2021.

\bibitem[Cohen and Welling(2016)]{cohen2016group}
T.~Cohen and M.~Welling.
\newblock Group equivariant convolutional networks.
\newblock In \emph{Proc. ICML}, 2016.

\bibitem[Cohen et~al.(2019)Cohen, Weiler, Kicanaoglu, and
  Welling]{cohen2019gauge}
T.~Cohen, M.~Weiler, B.~Kicanaoglu, and M.~Welling.
\newblock Gauge equivariant convolutional networks and the icosahedral {CNN}.
\newblock In \emph{Proc. ICML}, 2019.

\bibitem[Cohen et~al.(2018)Cohen, Geiger, Köhler, and
  Welling]{cohen2018spherical}
T.~S. Cohen, M.~Geiger, J.~Köhler, and M.~Welling.
\newblock Spherical {CNN}s.
\newblock In \emph{Proc. ICLR}, 2018.

\bibitem[Contributors(2020)]{mmseg2020}
M.~Contributors.
\newblock {MMSegmentation}: Openmmlab semantic segmentation toolbox and
  benchmark.
\newblock \url{https://github.com/open-mmlab/mmsegmentation}, 2020.

\bibitem[Cubuk et~al.(2020)Cubuk, Zoph, Shlens, and Le]{cubuk_2020_randaugment}
E.~D. Cubuk, B.~Zoph, J.~Shlens, and Q.~V. Le.
\newblock Randaugment: Practical automated data augmentation with a reduced
  search space.
\newblock In \emph{Proc. CVPR workshop}, 2020.

\bibitem[de~Haan et~al.(2021)de~Haan, Weiler, Cohen, and
  Welling]{dehaan2020gauge}
P.~de~Haan, M.~Weiler, T.~Cohen, and M.~Welling.
\newblock Gauge equivariant mesh {CNNs}: Anisotropic convolutions on geometric
  graphs.
\newblock In \emph{Proc. ICLR}, 2021.

\bibitem[Defferrard et~al.(2016)Defferrard, Bresson, and
  Vandergheynst]{defferrard2016convolutional}
M.~Defferrard, X.~Bresson, and P.~Vandergheynst.
\newblock Convolutional neural networks on graphs with fast localized spectral
  filtering.
\newblock In \emph{Proc. NeurIPS}, 2016.

\bibitem[Deng et~al.(2009)Deng, Dong, Socher, Li, Li, and
  Fei-Fei]{deng2009imagenet}
J.~Deng, W.~Dong, R.~Socher, L.-J. Li, K.~Li, and L.~Fei-Fei.
\newblock {ImageNet}: A large-scale hierarchical image database.
\newblock In \emph{Proc. CVPR}, 2009.

\bibitem[Ding et~al.(2023)Ding, Soselia, Armstrong, Su, and
  Huang]{ding_2023_reviving}
P.~Ding, D.~Soselia, T.~Armstrong, J.~Su, and F.~Huang.
\newblock Reviving shift equivariance in vision transformers.
\newblock \emph{arXiv preprint arXiv:2306.07470}, 2023.

\bibitem[Dosovitskiy et~al.(2021)Dosovitskiy, Beyer, Kolesnikov, Weissenborn,
  Zhai, Unterthiner, Dehghani, Minderer, Heigold, Gelly, Uszkoreit, and
  Houlsby]{dosovitskiy2020vit}
A.~Dosovitskiy, L.~Beyer, A.~Kolesnikov, D.~Weissenborn, X.~Zhai,
  T.~Unterthiner, M.~Dehghani, M.~Minderer, G.~Heigold, S.~Gelly, J.~Uszkoreit,
  and N.~Houlsby.
\newblock An image is worth 16x16 words: Transformers for image recognition at
  scale.
\newblock In \emph{Proc. ICLR}, 2021.

\bibitem[Fan et~al.(2021)Fan, Xiong, Mangalam, Li, Yan, Malik, and
  Feichtenhofer]{fan2021multiscale}
H.~Fan, B.~Xiong, K.~Mangalam, Y.~Li, Z.~Yan, J.~Malik, and C.~Feichtenhofer.
\newblock Multiscale vision transformers.
\newblock In \emph{Proc. CVPR}, 2021.

\bibitem[Han et~al.(2022)Han, Wang, Chen, Chen, Guo, Liu, Tang, Xiao, Xu, Xu,
  et~al.]{han2022survey}
K.~Han, Y.~Wang, H.~Chen, X.~Chen, J.~Guo, Z.~Liu, Y.~Tang, A.~Xiao, C.~Xu,
  Y.~Xu, et~al.
\newblock A survey on vision transformer.
\newblock \emph{IEEE TPAMI}, 2022.

\bibitem[Hartford et~al.(2018)Hartford, Graham, Leyton-Brown, and
  Ravanbakhsh]{hartford2018deep}
J.~Hartford, D.~Graham, K.~Leyton-Brown, and S.~Ravanbakhsh.
\newblock Deep models of interactions across sets.
\newblock In \emph{Proc. ICML}, 2018.

\bibitem[He et~al.(2016)He, Zhang, Ren, and Sun]{He_Zhang_Ren_Sun_2016}
K.~He, X.~Zhang, S.~Ren, and J.~Sun.
\newblock Deep residual learning for image recognition.
\newblock In \emph{Proc. CVPR}, 2016.

\bibitem[Khan et~al.(2022)Khan, Naseer, Hayat, Zamir, Khan, and
  Shah]{khan2022transformers}
S.~Khan, M.~Naseer, M.~Hayat, S.~W. Zamir, F.~S. Khan, and M.~Shah.
\newblock Transformers in vision: A survey.
\newblock \emph{ACM CSUR}, 2022.

\bibitem[Kipf and Welling(2017)]{kipf2017semi}
T.~N. Kipf and M.~Welling.
\newblock Semi-supervised classification with graph convolutional networks.
\newblock In \emph{Proc. ICLR}, 2017.

\bibitem[Klee et~al.(2023)Klee, Biza, Platt, and Walters]{klee2023image}
D.~M. Klee, O.~Biza, R.~Platt, and R.~Walters.
\newblock Image to sphere: Learning equivariant features for efficient pose
  prediction.
\newblock In \emph{Proc. ICLR}, 2023.

\bibitem[Kondor et~al.(2018)Kondor, Lin, and Trivedi]{kondor2018clebsch}
R.~Kondor, Z.~Lin, and S.~Trivedi.
\newblock {Clebsch\textendash Gordan Nets}: a fully {Fourier} space spherical
  convolutional neural network.
\newblock In \emph{Proc. NeurIPS}, 2018.

\bibitem[Krizhevsky et~al.(2009)Krizhevsky, Hinton,
  et~al.]{krizhevsky2009learning}
A.~Krizhevsky, G.~Hinton, et~al.
\newblock Learning multiple layers of features from tiny images.
\newblock 2009.

\bibitem[Krizhevsky et~al.(2012)Krizhevsky, Sutskever, and
  Hinton]{Krizhevsky_Sutskever_Hinton_2012}
A.~Krizhevsky, I.~Sutskever, and G.~E. Hinton.
\newblock {ImageNet} classification with deep convolutional neural networks.
\newblock In \emph{Proc. NeurIPS}, 2012.

\bibitem[Li et~al.(2022)Li, Wu, Fan, Mangalam, Xiong, Malik, and
  Feichtenhofer]{li2022mvitv2}
Y.~Li, C.-Y. Wu, H.~Fan, K.~Mangalam, B.~Xiong, J.~Malik, and C.~Feichtenhofer.
\newblock {MViTv2}: Improved multiscale vision transformers for classification
  and detection.
\newblock In \emph{Proc. CVPR}, 2022.

\bibitem[Lin et~al.(2013)Lin, Chen, and Yan]{lin2013network}
M.~Lin, Q.~Chen, and S.~Yan.
\newblock Network in network.
\newblock \emph{arXiv preprint arXiv:1312.4400}, 2013.

\bibitem[Liu et~al.(2020)Liu, Yeh, and Schwing]{liu2020pic}
I.-J. Liu, R.~A. Yeh, and A.~G. Schwing.
\newblock {PIC}: permutation invariant critic for multi-agent deep
  reinforcement learning.
\newblock In \emph{Proc. CORL}, 2020.

\bibitem[Liu et~al.(2021{\natexlab{a}})Liu, Ren, Yeh, and
  Schwing]{liu2021semantic}
I.-J. Liu, Z.~Ren, R.~A. Yeh, and A.~G. Schwing.
\newblock Semantic tracklets: An object-centric representation for visual
  multi-agent reinforcement learning.
\newblock In \emph{Proc. IROS}, 2021{\natexlab{a}}.

\bibitem[Liu et~al.(2021{\natexlab{b}})Liu, Lin, Cao, Hu, Wei, Zhang, Lin, and
  Guo]{liu_2021_swin}
Z.~Liu, Y.~Lin, Y.~Cao, H.~Hu, Y.~Wei, Z.~Zhang, S.~Lin, and B.~Guo.
\newblock Swin transformer: Hierarchical vision transformer using shifted
  windows.
\newblock In \emph{Proc. ICCV}, 2021{\natexlab{b}}.

\bibitem[Liu et~al.(2022)Liu, Hu, Lin, Yao, Xie, Wei, Ning, Cao, Zhang, Dong,
  et~al.]{liu2022swin}
Z.~Liu, H.~Hu, Y.~Lin, Z.~Yao, Z.~Xie, Y.~Wei, J.~Ning, Y.~Cao, Z.~Zhang,
  L.~Dong, et~al.
\newblock Swin transformer v2: Scaling up capacity and resolution.
\newblock In \emph{Proc. CVPR}, 2022.

\bibitem[Maron et~al.(2019)Maron, Ben-Hamu, Shamir, and
  Lipman]{maron2018invariant}
H.~Maron, H.~Ben-Hamu, N.~Shamir, and Y.~Lipman.
\newblock Invariant and equivariant graph networks.
\newblock In \emph{Proc. ICLR}, 2019.

\bibitem[Maron et~al.(2020)Maron, Litany, Chechik, and
  Fetaya]{maron2020learning}
H.~Maron, O.~Litany, G.~Chechik, and E.~Fetaya.
\newblock On learning sets of symmetric elements.
\newblock In \emph{Proc. ICML}, 2020.

\bibitem[Michaeli et~al.(2023)Michaeli, Michaeli, and
  Soudry]{michaeli_2023_alias}
H.~Michaeli, T.~Michaeli, and D.~Soudry.
\newblock Alias-free convnets: Fractional shift invariance via polynomial
  activations.
\newblock In \emph{Proc. CVPR}, 2023.

\bibitem[Morris et~al.(2022)Morris, Rattan, Kiefer, and Ravanbakhsh]{morris22a}
C.~Morris, G.~Rattan, S.~Kiefer, and S.~Ravanbakhsh.
\newblock {S}peq{N}ets: Sparsity-aware permutation-equivariant graph networks.
\newblock In K.~Chaudhuri, S.~Jegelka, L.~Song, C.~Szepesvari, G.~Niu, and
  S.~Sabato, editors, \emph{Proc. ICML}, 2022.

\bibitem[Qi et~al.(2017)Qi, Su, Mo, and Guibas]{qi2017pointnet}
C.~R. Qi, H.~Su, K.~Mo, and L.~J. Guibas.
\newblock {PointNet}: Deep learning on point sets for {3D} classification and
  segmentation.
\newblock In \emph{Proc. CVPR}, 2017.

\bibitem[Ravanbakhsh et~al.(2017{\natexlab{a}})Ravanbakhsh, Schneider, and
  P{\'o}czos]{ravanbakhsh2017equivariance}
S.~Ravanbakhsh, J.~Schneider, and B.~P{\'o}czos.
\newblock Equivariance through parameter-sharing.
\newblock In \emph{Proc. ICML}, 2017{\natexlab{a}}.

\bibitem[Ravanbakhsh et~al.(2017{\natexlab{b}})Ravanbakhsh, Schneider, and
  Poczos]{ravanbakhsh_sets}
S.~Ravanbakhsh, J.~Schneider, and B.~Poczos.
\newblock Deep learning with sets and point clouds.
\newblock In \emph{Proc. ICLR workshop}, 2017{\natexlab{b}}.

\bibitem[Rojas-Gomez et~al.(2022)Rojas-Gomez, Lim, Schwing, Do, and
  Yeh]{rojas-neurips2022-learnable}
R.~A. Rojas-Gomez, T.~Y. Lim, A.~G. Schwing, M.~N. Do, and R.~A. Yeh.
\newblock Learnable polyphase sampling for shift invariant and equivariant
  convolutional networks.
\newblock In \emph{Proc. NeurIPS}, 2022.

\bibitem[Romero et~al.(2020)Romero, Bekkers, Tomczak, and
  Hoogendoorn]{romero2020attentive}
D.~Romero, E.~Bekkers, J.~Tomczak, and M.~Hoogendoorn.
\newblock Attentive group equivariant convolutional networks.
\newblock In \emph{Proc. ICML}, 2020.

\bibitem[Romero and Lohit(2022)]{romero2022learning}
D.~W. Romero and S.~Lohit.
\newblock Learning partial equivariances from data.
\newblock In \emph{Proc. NeurIPS}, 2022.

\bibitem[Sandler et~al.(2018)Sandler, Howard, Zhu, Zhmoginov, and
  Chen]{Sandler_Howard_Zhu_Zhmoginov_Chen_2018}
M.~Sandler, A.~Howard, M.~Zhu, A.~Zhmoginov, and L.-C. Chen.
\newblock {MobileNetV2}: Inverted residuals and linear bottlenecks.
\newblock In \emph{Proc. CVPR}, 2018.

\bibitem[Shakerinava and Ravanbakhsh(2021)]{shakerinava21a}
M.~Shakerinava and S.~Ravanbakhsh.
\newblock Equivariant networks for pixelized spheres.
\newblock In \emph{Proc. ICML}, 2021.

\bibitem[Shuman et~al.(2013)Shuman, Narang, Frossard, Ortega, and
  Vandergheynst]{shuman2013emerging}
D.~I. Shuman, S.~K. Narang, P.~Frossard, A.~Ortega, and P.~Vandergheynst.
\newblock The emerging field of signal processing on graphs: Extending
  high-dimensional data analysis to networks and other irregular domains.
\newblock \emph{IEEE SPM}, 2013.

\bibitem[Simonyan and Zisserman(2015)]{Simonyan_Zisserman_2015}
K.~Simonyan and A.~Zisserman.
\newblock Very deep convolutional networks for large-scale image recognition.
\newblock In \emph{Proc. ICLR}, 2015.

\bibitem[van~der Ouderaa et~al.(2022)van~der Ouderaa, Romero, and van~der
  Wilk]{van2022relaxing}
T.~van~der Ouderaa, D.~W. Romero, and M.~van~der Wilk.
\newblock Relaxing equivariance constraints with non-stationary continuous
  filters.
\newblock In \emph{Proc. NeurIPS}, 2022.

\bibitem[Vaswani et~al.(2017)Vaswani, Shazeer, Parmar, Uszkoreit, Jones, Gomez,
  Kaiser, and Polosukhin]{vaswani_2017_attention}
A.~Vaswani, N.~Shazeer, N.~Parmar, J.~Uszkoreit, L.~Jones, A.~N. Gomez,
  {\L}.~Kaiser, and I.~Polosukhin.
\newblock Attention is all you need.
\newblock In \emph{Proc. NeurIPS}, 2017.

\bibitem[Venkataraman et~al.(2020)Venkataraman, Balasubramanian, and
  Sarma]{venkataraman2019building}
S.~R. Venkataraman, S.~Balasubramanian, and R.~R. Sarma.
\newblock Building deep equivariant capsule networks.
\newblock In \emph{Proc. ICLR}, 2020.

\bibitem[Vetterli et~al.(2014)Vetterli, Kova{\v{c}}evi{\'c}, and
  Goyal]{vetterli2014foundations}
M.~Vetterli, J.~Kova{\v{c}}evi{\'c}, and V.~K. Goyal.
\newblock \emph{Foundations of signal processing}.
\newblock Cambridge University Press, 2014.

\bibitem[Weiler and Cesa(2019)]{weiler2019general}
M.~Weiler and G.~Cesa.
\newblock General {E(2)}-equivariant steerable {CNNs}.
\newblock In \emph{Proc. NeurIPS}, 2019.

\bibitem[Wu et~al.(2021)Wu, Xiao, Codella, Liu, Dai, Yuan, and
  Zhang]{wu2021cvt}
H.~Wu, B.~Xiao, N.~Codella, M.~Liu, X.~Dai, L.~Yuan, and L.~Zhang.
\newblock {CvT}: Introducing convolutions to vision transformers.
\newblock In \emph{Proc. CVPR}, 2021.

\bibitem[Xiao et~al.(2018)Xiao, Liu, Zhou, Jiang, and Sun]{xiao2018unified}
T.~Xiao, Y.~Liu, B.~Zhou, Y.~Jiang, and J.~Sun.
\newblock Unified perceptual parsing for scene understanding.
\newblock In \emph{Proc. ECCV}, 2018.

\bibitem[Yeh et~al.(2019{\natexlab{a}})Yeh, Hu, and Schwing]{yeh2019chirality}
R.~A. Yeh, Y.-T. Hu, and A.~Schwing.
\newblock Chirality nets for human pose regression.
\newblock In \emph{Proc. NeurIPS}, 2019{\natexlab{a}}.

\bibitem[Yeh et~al.(2019{\natexlab{b}})Yeh, Schwing, Huang, and
  Murphy]{yeh2019diverse}
R.~A. Yeh, A.~G. Schwing, J.~Huang, and K.~Murphy.
\newblock Diverse generation for multi-agent sports games.
\newblock In \emph{Proc. CVPR}, 2019{\natexlab{b}}.

\bibitem[Yeh et~al.(2022)Yeh, Hu, Hasegawa-Johnson, and
  Schwing]{yeh2022equivariance}
R.~A. Yeh, Y.-T. Hu, M.~Hasegawa-Johnson, and A.~Schwing.
\newblock Equivariance discovery by learned parameter-sharing.
\newblock In \emph{Proc. AISTATS}, 2022.

\bibitem[Zaheer et~al.(2017)Zaheer, Kottur, Ravanbakhsh, Poczos, Salakhutdinov,
  and Smola]{zaheer2017deep}
M.~Zaheer, S.~Kottur, S.~Ravanbakhsh, B.~Poczos, R.~R. Salakhutdinov, and A.~J.
  Smola.
\newblock Deep sets.
\newblock In \emph{Proc. NeurIPS}, 2017.

\bibitem[Zhang(2019)]{zhang2019making}
R.~Zhang.
\newblock Making convolutional networks shift-invariant again.
\newblock In \emph{Proc. ICML}, 2019.

\bibitem[Zhou et~al.(2017)Zhou, Zhao, Puig, Fidler, Barriuso, and
  Torralba]{zhou2017scene}
B.~Zhou, H.~Zhao, X.~Puig, S.~Fidler, A.~Barriuso, and A.~Torralba.
\newblock Scene parsing through {ADE20K} dataset.
\newblock In \emph{Proc. CVPR}, 2017.

\bibitem[Zou et~al.(2020)Zou, Xiao, Yu, and Lee]{zou2020delving}
X.~Zou, F.~Xiao, Z.~Yu, and Y.~J. Lee.
\newblock Delving deeper into anti-aliasing in convnets.
\newblock In \emph{Proc. BMVC}, 2020.

\end{thebibliography}

\newpage
\appendix
\onecolumn
\setcounter{section}{0}
\renewcommand{\thesection}{A\arabic{section}}
\renewcommand{\thetable}{A\arabic{table}}
\setcounter{table}{0}
\setcounter{figure}{0}
\renewcommand{\thetable}{A\arabic{table}}
\renewcommand\thefigure{A\arabic{figure}}
\renewcommand{\theHtable}{A.Tab.\arabic{table}}
\renewcommand{\theHfigure}{A.Abb.\arabic{figure}}
\renewcommand\theequation{A\arabic{equation}}
\renewcommand{\theHequation}{A.Abb.\arabic{equation}}

{\centering \Large \textbf{Appendix}}\\

{\noindent}The appendix is organized as follows:
\begin{itemize}
\item \secref{supp_sec:proof} provides the complete proofs for all the claims stated in the main paper.
\item \secref{sec:supp_implt} includes additional implementation, runtime details, and memory consumption.
\item \secref{sec:supp_exp_details} fully describes the augmentation settings used for image classification and semantic segmentation.
\item \secref{sec:supp_results} provides additional semantic segmentation qualitative results.
\item \secref{sec:supp_additional_experiments} reports additional image classification results covering:
    \begin{enumerate}[label=\alph*.]
     \item The robustness of our proposed models to out-of-distribution images (\secref{sec:supp_ood}).
     \item The sensitivity of our ViT models to input shifts of different magnitudes (\secref{sec:supp_sensitivity}).
     \item The use of our proposed adaptive modules on pre-trained ViTs (\secref{sec:supp_pretrained}).
    \end{enumerate}
\end{itemize}

\section{Complete proofs}\label{supp_sec:proof}
\subsection{Proof of Lemma~\ref{clm:lemma}}

\begin{mdframed}[style=MyFrame]
\lem*
\end{mdframed}
\begin{proof}

By definition,
    \begin{align}
        \hat{\mX}^{(m)}=&\ \text{\texttt{reshape}}(\gS_{N}^{m}\hat{\vx})
        =\ \text{\texttt{reshape}}(\gS_{N}^{m+1}\vx)
    \end{align}
    Let the input patches be expressed as $\mX^{(m)}=\begin{bmatrix}\vr_{0}^{(m)} & \dots & \vr_{L-1}^{(m)}\end{bmatrix} \in \sR^{\floor{N/L}\times L}$, where $\vr_{k}^{(m)}\in \sR^{\floor{N/L}}$ is comprised by the $k^{\text{th}}$ element of every input patch.
    \begin{align}
        \vr_{k}^{(m)}[n]=&\ (\gS_{N}^{m}\vx)[Ln+k]
        =\ x[\circshift{(Ln+k+m)}{N}].
    \end{align}
    More precisely, given a circularly shifted input with offset $m$, $\vr_{k}^{(m)}[n]$ represents the $k^{\text{th}}$ element of the $n^{\text{th}}$ patch. %
    Based on this, $\hat{\mX}^{(m)}\in \sR^{\floor{N/L}\times L}$ can be expressed as:
    \begin{align}
        \hat{\mX}^{(m)}=\ \text{\texttt{reshape}}(\gS_{N}^{m+1}\vx)
        =&\ \begin{bmatrix} \vr_{0}^{(m+1)} & \dots & \vr_{L-1}^{(m+1)}\end{bmatrix},\\
        \text{with }
        \vr_{k}^{(m+1)}[n]=&\ \vx[\circshift{(Ln+k+m+1)}{N}].
    \end{align}
    Expressed in terms of its quotient and remainder for divisor $L$, $m+1=\ \floor{\frac{m+1}{L}}L+\circshift{(m+1)}{L}$. Then, $\hat{\vr}$ corresponds to:
    \begin{align}
        \hat{\vr}_{k}^{(m)}[n]=&\ \vx\left[\circshift{\left(L\left(n+\floor{\frac{m+1}{L}}\right)+\circshift{(m+1)}{L}+k\right)}{N}\right]
    \end{align}
    It follows that $\hat{\vr}_{k}^{(m)}=\ \gS_{\floor{\frac{N}{L}}}^{\floor{\frac{m+1}{L}}}\vr_{k}^{(\circshift{(m+1)}{L})}$, which implies
    \begin{align}
        \hat{\mX}^{(m)}\mE=&\ \gS_{\floor{\frac{N}{L}}}^{\floor{\frac{m+1}{L}}}\mX^{(\circshift{(m+1)}{L})}\mE.
    \end{align}
\end{proof}

\clearpage
\subsection{Proof of~\clmref{clm:a_token_shifteq}}
\begin{mdframed}[style=MyFrame]
\token*
\end{mdframed}
\begin{proof}
    Let $\hat{\vx}=\gS_{N} \vx \in \sR^{N}$ be a circularly shifted version of input $\vx\in \sR^{N}$. From Lemma \ref{clm:lemma}, their token representations satisfy:
    \begin{align}
        \label{eq:token_shifteq01}
        \hat{\mX}^{(m)}\mE=&\ \gS_{\floor{\frac{N}{L}}}^{\floor{\frac{m+1}{L}}}\mX^{(\circshift{(m+1)}{L})}\mE.
    \end{align}
    Based on the selection criterion of \texttt{A-token} and \equref{eq:token_shifteq01}:
    \begin{align}
        \underset{m\in\{0,\dots,L-1\}}{\max}\ F(\hat{\mX}^{(m)}\mE)=&\ \underset{m\in\{0,\dots,L-1\}}{\max}\ F\left(\gS_{\floor{\frac{N}{L}}}^{\floor{\frac{m+1}{L}}}\mX^{(\circshift{(m+1)}{L})}\mE\right)\\
        \label{eq:token_shifteq01_invariance}
        =&\ \underset{m\in\{0,\dots,L-1\}}{\max}\ F\left(\mX^{(\circshift{(m+1)}{L})}\mE\right),
    \end{align}
where the right-hand side in Eq. \eqref{eq:token_shifteq01_invariance} derives from the shift-invariance property of $F$. 
    Note that for any integer $m$ in the range $\{0,\dots,L-1\}$, the circular shift $\circshift{(m+1)}{L}$ also lies within the same range. It follows that:
    \begin{align}
        \label{eq:token_shifteq02}
        \underset{m\in\{0,\dots,L-1\}}{\max}\ F\left(\mX^{(\circshift{(m+1)}{L})}\mE\right)=&\ \underset{m\in\{0,\dots,L-1\}}{\max}\ F\big(\mX^{(m)}\mE\big).
    \end{align}
    Next, let $\hat{\mX}^{(\hat{m})}\mE=\ \text{\texttt{A-token}}(\hat{\vx})$. Then, from Lemma \ref{clm:lemma}:
    \begin{align}
        \label{eq:token_shifteq03}
        \underset{m\in\{0,\dots,L-1\}}{\max}\ F\big(\hat{\mX}^{(m)}\mE)=&\ F\big(\hat{\mX}^{(\hat{m})}\mE)
        =\ F\big(\mX^{(\circshift{(\hat{m}+1)}{L})}\mE)
    \end{align}
    Let $\mX^{(m^{\star})}\mE=\ \text{\texttt{A-token}}(\vx)$. Then, From~\equref{eq:token_shifteq02} and~\equref{eq:token_shifteq03}:
    \begin{align}
        \label{eq:token_shifteq04}
        \underset{m\in\{0,\dots,L-1\}}{\max}\ F\big(\mX^{(m)}\mE)=&\ F\big(\mX^{(\circshift{(\hat{m}+1)}{L})}\mE),
    \end{align}
    which implies that $m^{\star}=\circshift{(\hat{m}+1)}{L}$. Finally, from Lemma \ref{clm:lemma} and Eq. \eqref{eq:token_shifteq04}:
    \begin{align}
        \text{\texttt{A-token}}(\gS_{N}\vx)=&\ \gS_{\floor{\frac{N}{L}}}^{\floor{\frac{(\hat{m}+1)}{L}}}\mX^{(\circshift{(\hat{m}+1)}{L})}\mE
        =\ \gS_{\floor{\frac{N}{L}}}^{\floor{\frac{(\hat{m}+1)}{L}}}\text{\texttt{A-token}}(\vx).
    \end{align}
\end{proof}

\subsection{Proof of~\clmref{claim:awsa_equivariance}} 
\begin{mdframed}[style=MyFrame]
\awsa*
\end{mdframed}
\begin{proof}
    Let $\mT \in \sR^{M}$ and $\hat{\mT}=\gS_M\mT\in \sR^{M}$ denote two token representations related by a circular shift. From the definition of $\vv_{W}^{(m)}$ in \equref{eq:awsa_energy_shift0}, let $\vv \in \mathbb{R}^{M}$ denote the average $\ell_{p}$-norm (energy) of each group of $W$ neighboring tokens in $\mT$. More precisely, the $k$-th component of the energy vector $\vv$, denoted as $\vv[k]$, is the energy of the window comprised by $W$ neighboring tokens starting at index $k$.
    \begin{align}
        \label{eq:awsa_energy_vector}
        \vv[k]=&\ \frac{1}{W}\sum_{l=0}^{W-1}\|\mT_{\circshift{(k+l)}{M}}\|_{p}
    \end{align}
    Similarly, following Eq. \eqref{eq:awsa_energy_shift}, let $\vv_{W}^{(m)}\in \sR^{\floor{\frac{M}{W}}}$ and $\hat{\vv}_{W}^{(m)}\in \sR^{\floor{\frac{M}{W}}}$ denote the energy vectors of non-overlapping windows obtained after shifting $\mT$ and $\hat{\mT}$ by $m$ indices, respectively. Then, maximizers $m^{\star}$ and $\hat{m}$ satisfy
    \begin{align}
        \label{eq:awsa_maximizers}
        m^{\star}=&\ \argmax_{m\in\{0,\dots,W-1\}}G\big(\vv_{W}^{(m)}\big),\
        \hat{m}=\ \argmax_{m\in\{0,\dots,W-1\}}G\big(\hat{\vv}_{W}^{(m)}\big)
    \end{align}
    In what follows, we prove that their adaptive window-based self-attention outputs are related by a circular shift that is a multiple of the window size $W$, \ie, there exists $m_{0}\in \sZ$ such that
    \begin{align}
        \text{\texttt{A-WSA}}(\gS_{M}\mT)=\gS_{M}^{m_{0}W}\text{\texttt{A-WSA}}(\mT).
    \end{align}
    Given the shifted input token representation $\hat{\mT}=\gS_{M}\mT$, the energy of each group of $W$ neighboring tokens corresponds to
    \begin{align}
        \hat{\vv}[k]=&\ \frac{1}{W}\sum_{l=0}^{W-1}\|(\gS_{M}\mT)_{\circshift{(k+l)}{M}}\|_{p}\ = \frac{1}{W}\sum_{l=0}^{W-1}\|\mT_{\circshift{(k+l+1)}{M}}\|_{p}\\
        =&\ \vv[k+1]
    \end{align}
    which implies $\hat{\vv}=\gS_{M}\vv$. Then, $\hat{\vv}_{W}^{(m)}$ can be expressed as
    \begin{align}
        \hat{\vv}_{W}^{(m)}[k]=&\ \hat{\vv}[Wk+m]
        =\ \vv[Wk+m+1].
    \end{align}
    Expressing $m+1$ in terms of its quotient and remainder for divisor $W$
    \begin{align}
        \label{window_shifteq}
        \hat{\vv}_{M}^{(m)}[k]=&\ \vv\bigg[W\big(k+\floor{\frac{m+1}{W}}\big)+(m+1)\ \text{\texttt{mod}}\ W\bigg]\\
        =&\ \gS_{\floor{\frac{M}{W}}}^{\floor{\frac{m+1}{W}}}\vv_{W}^{((m+1)\ \text{\texttt{mod}}\ W)}.
    \end{align}
    Based on the shift-invariant property of $G$ and the fact that $(m+1)\ \text{\texttt{mod}}\ W \in \{0,\dots,W-1\}$, \texttt{A-WSA} selection criterion corresponds to
    \begin{align}
        \max_{m\in \{0,\dots,W-1\}} G\big(\hat{\vv}_{W}^{(m)}\big)=&\ \max_{m\in \{0,\dots,W-1\}} G\big(\gS_{\floor{\frac{M}{W}}}^{\floor{\frac{m+1}{W}}}\vv_{W}^{((m+1)\ \text{\texttt{mod}}\ W)}\big)\\
        =&\ \max_{m\in \{0,\dots,W-1\}} G\big(\vv_{W}^{((m+1)\ \text{\texttt{mod}}\ W)}\big)\\
        =&\ \max_{m\in \{0,\dots,W-1\}} G\big(\vv_{W}^{(m)}\big).
    \end{align}
    Then, from $\hat{m}$ in \equref{eq:awsa_maximizers}
    \begin{align}
        \max_{m\in \{0,\dots,W-1\}} G\big(\hat{\vv}_{W}^{(m)}\big)=&\ G\big(\hat{\vv}_{W}^{(\hat{m})}\big)=\ G\left(\vv_{W}^{((\hat{m}+1)\ \text{\texttt{mod}}\ W)}\right)
    \end{align}
    which implies $m^{\star}= (\hat{m}+1)\ \text{\texttt{mod}}\ W$.
    It follows that the adaptive self-attention of $\hat{\mT}$ can be expressed as:
    \begin{align}
        \text{\texttt{A-WSA}}(\hat{\mT})=&\ \text{\texttt{WSA}}(\gS_{M}^{\hat{m}}\hat{\mT})\
        =\ \text{\texttt{WSA}}(\gS_{M}^{\hat{m}+1}\mT)\\
        =&\ \text{\texttt{WSA}}\left(\gS_{M}^{\floor{(\hat{m}+1)/W}W+m^{\star}}\mT\right)\\
        =&\ \text{\texttt{WSA}}\left(\gS_{M}^{\floor{(\hat{m}+1)/W}W} \gS_{M}^{m^{\star}}\mT\right).
    \end{align}
    Since $\gS_{M}^{\floor{(\hat{m}+1)/W}W}$ corresponds to a circular shift by a multiple of the window size $W$
    \begin{align}
        \label{eq:window_shifteq02}
        \text{\texttt{A-WSA}}(\hat{\mT})=&\ \gS_{M}^{\floor{(\hat{m}+1)/W}W} \text{\texttt{WSA}}(\gS_{M}^{m^{\star}}\mT)
    \end{align}
    where the right-hand side of Eq. \eqref{eq:window_shifteq02} stems from the fact that, for \texttt{WSA} with a window size $W$, circularly shifting the input tokens by a multiple of $W$ results in an identical circular shift of the output tokens. Finally, from the definition of adaptive self-attention in \equref{eq:a_wsa}
    \begin{align}
        \text{\texttt{A-WSA}}(\hat{\mT})=&\ \gS_{M}^{m_{0}W}\text{\texttt{A-WSA}}(\mT),\ m_{0}=\floor{(\hat{m}+1)/W}.
    \end{align}
\end{proof}

Claim~\ref{claim:awsa_equivariance} shows that \texttt{A-WSA} induces an offset between token representations that is a multiple of the window size $W$. As a result, windows are comprised by the same tokens, despite circular shifts. This way, \texttt{A-WSA} guarantees that an input token representation and its circularly shifted version are split into the same token windows, leading to a circularly shifted self-attention output.

\subsection{Proof of~\clmref{claim:pmergeconv}} 
\begin{mdframed}[style=MyFrame]
\pmergeconv*
\end{mdframed}
\begin{proof}
Let the input tokens be expressed as $\mT= \begin{bmatrix}\vt_{0} & \dots & \vt_{D-1}\end{bmatrix}\in \sR^{M\times D}$, where $\vt_{j}\in \sR^{M}$ is comprised by the $j^{\text{th}}$ element of every input token. This implies that, given the $l^{\text{th}}$ token $\mT_{l}\in \sR^{D}$, $\mT_{l}[j]=\vt_{j}[l]$. Then, the patch merging output $\mZ= \begin{bmatrix}\vz_{0} & \dots & \vz_{\tilde{D}-1}\end{bmatrix}$ can be expressed as
\begin{align}
   \mZ=&\ \text{\texttt{PMerge}}(\mT)=\ \gD^{(P)}(\mY)\in \sR^{\frac{M}{P}\times \tilde{D}},
\end{align}
where $\gD^{(P)}\in \sR^{\frac{M}{P}\times M}$ is a downsampling operator of factor $P$, and $\mY=\begin{bmatrix}\vy_{0} & \dots & \vy_{\tilde{D}-1}\end{bmatrix}\in \sR^{M\times \tilde{D}}$ is the output of a convolution sum with kernel size $P$ and $\tilde{D}$ output channels
\begin{align}
    \label{eq:pmerge_conv}
   \vy_{k} =&\ \sum_{j=0}^{D-1}\vt_{j}\circledast \vh^{(k,j)}\in \sR^{M}
\end{align}
where $\circledast$ denotes circular convolution and $\vh^{(k,j)}=\begin{bmatrix}\vh^{(k,j)}_{0} & \dots & \vh^{(k,j)}_{P-1}\end{bmatrix}^{\top}\in \sR^{P}$ denotes a convolutional kernel. Note that the convolution sum in \equref{eq:pmerge_conv} involves $D\cdot \tilde{D}$ kernels, given that $k\in\{0,\dots,\tilde{D}-1\}$ and $j\in\{0,\dots,D-1\}$. Without loss of generality, assume the number of input tokens $M$ is divisible by the patch length $P$. Then, due to the linearity of the downsampling operator, $\vz_{k}\in \sR^{\frac{M}{P}}$ corresponds to
\begin{align}
    \label{eq:pmerge_conv02}
   \vz_{k} =&\ \sum_{j=0}^{D-1}\gD^{(P)}\left(\vt_{j}\circledast \vh^{(k,j)}\right).
\end{align}
Let  $\mH^{(k,j)}\in \sR^{M\times M}$ be a convolutional matrix representing kernel $\vh^{(k,j)}$
\begin{align}
    \mH^{(k,j)}=&\ \begin{bmatrix}
        \vh^{(k,j)}_{0} & \cdots & \vh^{(k,j)}_{P-1}\\
        & \vh^{(k,j)}_{0} & \cdots & \vh^{(k,j)}_{P-1}\\
        & & \ddots &\\
        \cdots & \vh^{(k,j)}_{P-1} & \cdots & \vh^{(k,j)}_{0}
    \end{bmatrix}.
\end{align}
Then, note that each summation term in \equref{eq:pmerge_conv02} can be expressed as a matrix-vector multiplication
\begin{align}
    \label{eq:pmerge_conv03}
    \vz_{k}=&\ \sum_{j=0}^{D-1}\tilde{\mH}^{(k,j)}\vt_{j}
\end{align}
where $\tilde{\mH}^{(k,j)}=\gD^{(P)}\mH^{(k,j)}\in \sR^{\frac{M}{P}\times M}$ corresponds to the convolutional matrix $\mH^{(k,j)}$ downsampled by a factor $P$ along the row index
\begin{align}
    \tilde{\mH}^{(k,j)}=&\ \begin{bmatrix}
        \vh^{(k,j)}_{0} & \cdots & \vh^{(k,j)}_{P-1}\\
        & & & \vh^{(k,j)}_{0} & \cdots & \vh^{(k,j)}_{P-1}\\
        & & & & & & \ddots \\
        & & & & & & & \vh^{(k,j)}_{0} & \cdots & \vh^{(k,j)}_{P-1}
    \end{bmatrix}.
\end{align}
Based on this, the convolution sum in \equref{eq:pmerge_conv02} can be expressed in matrix-vector form as
\begin{align}
    \label{eq:pmerge_conv04}
    \vz_{k}=&\ \begin{bmatrix}
        \tilde{\mH}^{(k,0)} & \cdots & \tilde{\mH}^{(k,D-1)}
    \end{bmatrix}\begin{bmatrix}
    \vt_{0} \\ \vdots \\ \vt_{D-1}
    \end{bmatrix}.
\end{align}
Then, from the patch representation of $\mT$ in \equref{eq_patch_merge01}, $\vz_{k}$ can be alternatively expressed as
\begin{align}
    \vz_{k}=&\ \tilde{\mT} \tilde{\ve}_{k}
\end{align}
where $\tilde{\ve}_{k}$ is the $k^{\text{th}}$ column of $\tilde{\mE}=\begin{bmatrix} \tilde{\ve}_{0} & \cdots & \tilde{\ve}_{\tilde{D}-1} \end{bmatrix}\in \sR^{PD \times \tilde{D}}$. Let $\texttt{vec}(\bar{\mT}_{P}^{(k)})\in \sR^{PD}$ operate in a column-wise manner:
\begin{align}
    \text{\texttt{vec}}(\bar{\mT}_{P}^{(k)})=&\ \text{\texttt{vec}}\big(\begin{bmatrix} \mT_{Pk} & \dots & \mT_{P(k+1)-1} \end{bmatrix}^{\top}\big)\\
    =&\ \begin{bmatrix}\mT_{Pk}[0] & \cdots & \mT_{P(k+1)-1}[0] & \mT_{Pk}[1] & \cdots & \mT_{P(k+1)-1}[1] & \cdots & \mT_{P(k+1)-1}[D-1]\end{bmatrix}^{\top}\\
    =&\ \begin{bmatrix}\vt_{0}[Pk] & \cdots & \vt_{0}[P(k+1)-1] & \vt_{1}[Pk] & \cdots & \vt_{1}[P(k+1)-1] & \cdots & \vt_{D-1}[P(k+1)-1]\end{bmatrix}^{\top}.
\end{align}
Finally, from \equref{eq:pmerge_conv04}, $\tilde{\ve}_{k}$ is equivalent to
\begin{align}
    \tilde{\ve}_{k}=&\ \begin{bmatrix}\vh^{(k,0)}\ ;\ \dots\ ;\ \vh^{(k,D-1)}\end{bmatrix}\in \sR^{DP}
\end{align}
\end{proof}

\clmref{claim:pmergeconv} shows that the original patch merging function \texttt{PMerge} is equivalent to projecting all $M$ overlapping patches of length $P$, which can be expressed as a circular convolution, followed by keeping only $\frac{M}{P}$ of the resulting tokens. Note that the selection of such tokens is done via a downsampling operation of factor $P$, which is not the only way of selecting them. In fact, there are $P$ different token representations that can be selected, as explained in \secref{sec:app}.

Based on this observation, the proposed \texttt{A-PMerge} introduced in \secref{sec:app} takes advantage of the polyphase decomposition to select the token representation in a data-dependent fashion, leading to a shift-equivariant patch merging scheme.

\section{Additional implementation details}\label{sec:supp_implt}
We have attached our code in the supplementary materials. We now provide an illustration of how to use our implementation and verify that the proposed modules are indeed truly shift-equivariant.
\subsection{\texttt{A-pmerge} usage}
We show a toy example of how to use the adaptive patch merging layer (\texttt{A-pmerge}) by building a simple image classifier. The model is comprised by an \texttt{A-pmerge} layer, which includes a linear embedding, followed by a global average pooling layer and finally a linear classification head. We empirically show that this simple model is shift-invariant.
\begin{minted}[bgcolor=mygr]{python}
# A-pmerge based classifier
class ApmergeClassifier(nn.Module):
  def __init__(
    self,stride,input_resolution,
    dim,num_classes=4,conv_padding_mode='circular'
  ):
    super().__init__()
    self.stride=stride
    # Pooling layer for A-pmerge
    pool_layer = partial(
      PolyphaseInvariantDown2D,
      component_selection=max_p_norm,
      antialias_layer=None,
    )
    # A-pmerge
    # No adaptive window selection
    # for illustration purposes
    self.apmerge = AdaptivePatchMerging(
      input_resolution=input_resolution,
      dim=dim,
      pool_layer=pool_layer,
      conv_padding_mode=conv_padding_mode,
      stride=stride,
      window_selection=None,
      window_size=None,
    )
    # Global pooling and head
    self.avgpool=nn.AdaptiveAvgPool2d((1,1))
    self.fc=nn.Linear(dim*stride,num_classes)

  def forward(self,x):
    # Reshape
    B,C,H,W = x.shape
    x = x.permute(0,2,3,1).reshape(B,H*W,C)
    # Adaptive patch merge
    x = self.apmerge(x)
    # Reshape back
    x = x.reshape(
      B,H//self.stride,W//self.stride,C*self.stride,
    ).permute(0,3,1,2)
    # Global average pooling
    x = torch.flatten(self.avgpool(x),1)
    # Classification head
    x = self.fc(x)
    return x

# Input tokens
B,C,H,W = 1,3,8,8
stride = 2
x = torch.randn(B,C,H,W).cuda().double()
# Shifted input
shift = torch.randint(-3,3,(2,))
x_shift = torch.roll(input=x,dims=(2,3),shifts=(shift[0],shift[1]))
# A-pmerge classifier
model = ApmergeClassifier(
  stride=stride,
  input_resolution=(H,W),
  dim=C).cuda().double().eval()
# Predict
y = model(x)
y_shift = model(x_shift)
err = torch.norm(y-y_shift)
assert(torch.allclose(y,y_shift))
# Check circularly shift invariance
print('y: {}'.format(y))
print('y_shift: {}'.format(y_shift))
print("error: {}".format(err))
\end{minted}
{\noindent \bf Out:}
\begin{mdframed}[backgroundcolor=myoo,  linecolor=white]
\begin{lstlisting}
y: tensor([[-0.1878,  0.2348,  0.0982, -0.2191]],
    device='cuda:0', dtype=torch.float64,
    grad_fn=<AddmmBackward0>)
y_shift: tensor([[-0.1878,  0.2348,  0.0982, -0.2191]],
    device='cuda:0', dtype=torch.float64,
    grad_fn=<AddmmBackward0>)
error: 0.0
\end{lstlisting}
\end{mdframed}

\subsection{Simple encoder-decoder usage}
To illustrate the use of our adaptive layers for semantic segmentation purposes, we demonstrate the use of the \texttt{A-PMerge}'s optimal indices and the \texttt{A-WSA}'s optimal offset for unpooling purposes. We build a simple encoder-decoder based on an \texttt{A-PMerge} layer equipped with \texttt{A-WSA} (implemented as \texttt{poly\_win} in the example code) to encode input tokens, followed by an unpooling module to place features back into their original positions at high resolution.

This example depicts the strategy used by our proposed semantic segmentation models, where backbone indices are passed to the segmentation head to upscale feature maps in a consistent fashion, leading to a shift-equivariant model.
\begin{minted}[bgcolor=mygr]{python}
# Simple encoder-decoder
class EncDec(nn.Module):
  def __init__(self,dims,stride,win_sel,win_sz,pad):
    super().__init__()
    self.pad = pad
    self.stride = stride
    self.win_sz = win_sz

    # A-pmerge equipped with A-WSA
    pool_layer = get_pool_method(name='max_2_norm',antialias_mode='skip')
    self.apmerge = AdaptivePatchMerging(
      dim=dims,
      pool_layer=pool_layer,
      conv_padding_mode=pad,
      stride=stride,
      window_size=win_sz,
      window_selection=win_sel,
    )
    # Unpool
    unpool_layer = get_unpool_method(unpool=True,pool_method='max_2_norm',
        antialias_mode='skip')
    self.unpool = set_unpool(unpool_layer=unpool_layer,stride=stride,
        p_ch=dims)

  def forward(self,x,hw_shape):
    B,C,H,W = x.shape
    # Reshape
    x = x.permute(0,2,3,1).reshape(B,H*W,C)
    # A-pmerge + A-WSA
    x_w,hw_shape,idx = self.apmerge(x=x,hw_shape=hw_shape,ret_indices=True)
    # Keep merging and windowing indices
    idx = idx[:3]
    # Reshape back
    x_w = x_w.reshape(
      B,H//self.stride,W//self.stride,C*self.stride).permute(0,3,1,2)
    #  Unroll and Unpool
    y = unroll_unpool_multistage(
      x=x_w,
      scale_factor=[self.stride],
      unpool_layer=self.unpool,
      idx=[idx],
      unpool_winsel_roll=self.pad,
    )
    return y

# Input
B,C,H,W = 1,3,28,28
shift_max = 6
x = torch.randn(B,C,H,W).cuda().double()
# Offsets
s01 = torch.randint(low=-shift_max,high=shift_max,size=(1,2)).tolist()[0]
s02 = torch.randint(low=-shift_max,high=shift_max,size=(1,2)).tolist()[0]
s03 = [s02[0]-s01[0],s02[1]-s01[1]]
# Shifted inputs
x01 = torch.roll(x,shifts=s01,dims=(-1,-2))
x02 = torch.roll(x,shifts=s02,dims=(-1,-2))
# Build encoder-decoder
# Use A-WSA (poly_win)
model = EncDec(dims=3,stride=2,pad='circular',win_sz=7,win_sel=poly_win,
    ).cuda().double().eval()
# Predictions
y01 = model(x01,hw_shape=(H,W))
y02 = model(x02,hw_shape=(H,W))
# Shift to compare
z = torch.roll(y01,shifts=s03,dims=(-1,-2))
err = torch.norm(z-y02)
assert torch.allclose(z,y02)
print("torch.norm(z-y02): {}".format(err))
\end{minted}
{\noindent \bf Out:}
\vspace{0.5 cm}
\begin{mdframed}[backgroundcolor=myoo,linecolor=white]
\begin{lstlisting}
torch.norm(z-y02): 0.0
\end{lstlisting}
\end{mdframed}

\subsection{Memory Consumption}
We compare the computational requirements of our adaptive Vision Transformer models and their default versions in terms of training and inference GPU memory. We report memory consumption on a single NVIDIA Quadro RTX 5000 GPU, where both training and inference are performed using batch size $64$ and the default image size per model.

Following previous work \cite{liu_2021_swin, liu2022swin, fan2021multiscale}, we compare the \textit{allocated} GPU memory required by each model via NVIDIA's \texttt{torch.cuda.max\_memory\_allocated()} command. This measures the maximum GPU memory occupied by tensors since the beginning of the executed program. Additionally, we report the maximum \textit{reserved} GPU memory, as assigned by the CUDA memory allocator, using the \texttt{torch.cuda.max\_memory\_reserved()} command. This measures the maximum memory occupied by tensors plus the extra memory reserved to speed up allocations. The required memory is reported in Megabytes ($1$ MiB $=2^{20}$ bytes). We also report the relative change, which corresponds to the memory increase with respect to the default models.

\tabref{tab:supp_memory} shows the memory required for training and inference for each of our adaptive models and their default versions. In terms of training, our adaptive framework marginally increases the allocated memory requirements up to $16\%$ for MViTv2, while the rest of the models increase the memory requirements by less than $8 \%$. This trend is also seen in the reserved memory, where it increases up to $15 \%$.
Additionally, in terms of inference, only our adaptive MViTv2 model marginally increases the allocated memory by $11\%$, while the rest of our adaptive models do not increase the required allocated memory. In terms of reserved memory, our adaptive models increase the requirement at most by $2 \%$, while in some cases the reserved memory decreases by $3 \%$.

Overall, the training memory consumption marginally increases with respect to the default models, while the inference memory consumption remains almost unaffected. 

\section{Additional experimental details}
\begin{table}[t]
\centering
\small
\setlength{\tabcolsep}{3.5pt}
\rowcolors{1}{}{bg_blue}
\resizebox{\columnwidth}{!}{%
\begin{tabular}{ccc|cc}
\toprule
\hiderowcolors 
\multirow{2}{*}{Model} & \multicolumn{2}{c|}{Training} & \multicolumn{2}{c}{Inference}\\
& \makecell{Max. Allocated Memory (MiB)} & \makecell{Max. Reserved Memory (MiB)} & \makecell{Max. Allocated Memory (MiB)} & \makecell{Max. Reserved Memory (MiB)}\\
\midrule
\showrowcolors
Swin-T & $4,935$ & $5,380$ & $1,012$ & $1,352$\\
A-Swin-T \textbf{(Ours)} & $5,178$ ($+4.92 \%$) & $5,678$ ($+5.54 \%$)& $1,012$ ($0 \%$)& $1,382$ ($+2.22 \%$)\\
SwinV2-T & $7,712$ & $8,016$ & $1,275$ & $1,544$\\
A-SwinV2-T \textbf{(Ours)}  & $8,113$  ($+5.2 \%$) & $8,710$ ($+8.66 \%$)& $1,275$ ($0 \%$)& $1,534$ ($-0.65 \%$)\\
CvT-13  & $5,794$ & $6,044$ & $1,587$ & $1,850$\\
A-CvT-13 \textbf{(Ours)}  & $6,257$  ($+7.99 \%$) & $6,528$ ($+8.01 \%$)& $1,587$ ($0 \%$)& $1,780$ ($-3.78 \%$)\\
MViTv2-T  & $6,335$ & $7,352$ & $1,944$ & $3,356$\\
A-MViTv2-T \textbf{(Ours)}  & $7,364$  ($+16.24 \%$) & $8,456$ ($+15.02 \%$)& $2,164$ ($+11.32 \%$)& $3,378$ ($+0.67 \%$)\\
\bottomrule
\end{tabular}
}
\vspace{-0.2cm}
\caption{\textbf{Memory Consumption:} Maximum allocated and reserved memory required by our adaptive ViTs and their default versions. Training and inference memory consumption is calculated on a single NVIDIA Quadro RTX 5000 GPU (batch size $64$, default image size per model),  and reported in Megabytes (MiB). We also report the relative change with respect to the default models $(\%)$, which is shown in parentheses.}
\label{tab:supp_memory}
\end{table}

\label{sec:supp_exp_details}

\subsection{Image classification}

{\bf\noindent Data pre-processing (ImageNet circular shifts, CIFAR10/100).}
Similarly to previous work on shift-equivariant CNNs \citep{chaman2021truly,rojas-neurips2022-learnable}, to highlight the fact that perfect shift equivariance is imposed by design and not induced during training, no shift or resizing augmentation is applied.~\tabref{tab:supp_class_aug_circular} includes the data pre-processing used in our ImageNet experiments under circular shifts, and in all our CIFAR10/100 experiments. As explained in~\secref{sec:classification}, input images are transformed to match the default image size used by each model. This process is denoted as \textit{Size pre-processing} in the table.
\begin{table}[t]
\begin{center}
\small
\begin{tabular}{c|c|c}
\specialrule{.15em}{.05em}{.05em} 
& Train Set & Test Set\\
\midrule
\makecell{Size Preprocessing} & \multicolumn{2}{c}{\makecell{(i) Resize (Swin, MViTv2, CvT: $256 \times 256$. SwinV2: $292 \times 292$)\\ (ii) Center Crop (Swin, MViTv2, CvT: $224 \times 224$, SwinV2: $256 \times 256$)}}\\
\specialrule{.15em}{.05em}{.05em}
\makecell{Augmentation} & \makecell{(i) Random Horizontal Flipping \\ (ii) Normalization} & \makecell{Normalization}\\
\specialrule{.15em}{.05em}{.05em}
\end{tabular}
\vspace{-0.25 cm}
\caption{\label{tab:supp_class_aug_circular} \textbf{Image Classification preprocessing (circular shifts).} Data preprocessing and augmentation used to train and test all default and adaptive ViTs under circular shifts.}
\end{center}
\vspace{-0.5 cm}
\end{table}

{\bf\noindent Data pre-processing (ImageNet standard shifts).}
In contrast to the circular shift scenario, each model uses its default data augmentation configuration for the ImageNet standard shift scenario. Tables \ref{tab:supp_class_aug_standard_swin}, \ref{tab:supp_class_aug_standard_mvit} and \ref{tab:supp_class_aug_standard_cvt} include the augmentation details of each model, A-Swin, A-SwinV2, A-MViTv2, and A-CVT respectively. Similarly to the circular shift case, all models use a data pre-processing step to change the input image size.

Note that all four models use RandAugment~\cite{cubuk_2020_randaugment} as data augmentation pipeline, which includes: \texttt{AutoContrast}, \texttt{Equalize}, \texttt{Invert}, \texttt{Rotate}, \texttt{Posterize}, \texttt{Solarize}, \texttt{ColorIncreasing}, \texttt{ContrastIncreasing}, \texttt{BrightnessIncreasing}, \texttt{SharpnessIncreasing}, \texttt{Shear-x}, \texttt{Shear-y}, \texttt{Translate-x}, and \texttt{Translate-y}.

\begin{table}[ht]
\begin{center}
\small
\begin{tabular}{c|c|c}
\specialrule{.15em}{.05em}{.05em} 
& Train Set & Test Set\\
\midrule
\makecell{Size Preprocessing} & \multicolumn{2}{c}{\makecell{(i) Resize (Swin: $256 \times 256$, SwinV2: $292 \times 292$)\\ (ii) Center Crop (Swin: $224 \times 224$, SwinV2: $256 \times 256$)}}\\
\midrule
\makecell{Augmentation} & \makecell{(i) Random Horizontal Flipping\\ (ii) RandAugment\\ (magnitude: $9$\\ increase severity: True,\\ augmentations per image: $2$,\\ standard deviation: $0.5$)\\ (iii) Normalization\\ (iv) RandomErasing 
} & \makecell{Normalization}\\
\specialrule{.15em}{.05em}{.05em}
\end{tabular}
\end{center}
\vspace{-0.5 cm}
\caption{\label{tab:supp_class_aug_standard_swin} \textbf{Swin / SwinV2 Image Classification preprocessing (standard shifts).} Data preprocessing and augmentation used to train and test the default (Swin, SwinV2) and adaptive (A-Swin, A-SwinV2) models.}

\end{table}

\begin{table}[ht]
\begin{center}
\small
\begin{tabular}{c|c|c}
\specialrule{.15em}{.05em}{.05em} 
& Train Set & Test Set\\
\midrule
\makecell{Size Preprocessing} & \multicolumn{2}{c}{\makecell{(i) Resize ($256 \times 256$)\\ (ii) Center Crop ($224 \times 224$)}}\\
\midrule
\makecell{Augmentation} & \makecell{(i) Random Horizontal Flipping\\ (ii) RandAugment\\ (magnitude: $10$\\ increase severity: True,\\ augmentations per image: $6$,\\ standard deviation: $0.5$)\\ (iii) Normalization\\ (iv) RandomErasing 
} & \makecell{Normalization}\\
\specialrule{.15em}{.05em}{.05em}
\end{tabular}
\end{center}
\vspace{-0.5 cm}
\caption{\label{tab:supp_class_aug_standard_mvit} \textbf{MViTv2 Image Classification preprocessing (standard shifts).} Data preprocessing and augmentation used to train and test the default (MViTv2) and adaptive (A-MViTv2) models.}
\end{table}
\begin{table}[ht]
\begin{center}
\small
\begin{tabular}{c|c|c}
\specialrule{.15em}{.05em}{.05em} 
& Train Set & Test Set\\
\midrule
\makecell{Size Preprocessing} & \multicolumn{2}{c}{\makecell{(i) Resize ($256 \times 256$)\\ (ii) Center Crop ($224 \times 224$)}}\\
\midrule
\makecell{Augmentation} & \makecell{(i) Random Horizontal Flipping\\ (ii) RandAugment\\ (magnitude: $9$\\ increase severity: True,\\ augmentations per image: $2$,\\ standard deviation: $0.5$)\\ (iii) Normalization\\ (iv) RandomErasing 
} & \makecell{Normalization}\\
\specialrule{.15em}{.05em}{.05em}
\end{tabular}
\end{center}
\vspace{-0.5 cm}
\caption{\label{tab:supp_class_aug_standard_cvt} \textbf{CvT Image Classification preprocessing (standard shifts).} Data preprocessing and augmentation used to train and test the default (CvT) and adaptive (A-CvT) models.}
\end{table}

{\bf\noindent Computational settings.}
CIFAR10/100 classification experiments using all four models were trained on two NVIDIA Quadro RTX 5000 GPUs with a batch size of $48$ images for $100$ epochs. On the other hand, ImageNet classification experiments were trained on eight NVIDIA A100 GPUs, where all models used their default effective batch size and numerical precision for $300$ epochs.

\subsection{Semantic Segmentation}
{\bf\noindent Data pre-processing (Circular shifts).}
For Swin+UperNet, both the baseline and adaptive (A-Swin) models were trained by resizing input images to size $1,792 \times 448$ followed by a random cropping of size $448 \times 448$. Next, the default data augmentation used in the Swin + UperNet model is used. During testing, input images are resized to $1792 \times 448$. Then, each dimension size is rounded up to the next multiple of $224$.~\tabref{tab:supp_seg_aug_circular_swin} describes the pre-processing and data augmentation pipelines.

For Swin-V2+UperNet, both the baseline and the adaptive (A-SwinV2) models were trained by resizing input images to size $2,048\times 512$ followed by a random cropping of size $512\times 512$. Following this, the default data augmentation used in the Swin + UperNet model is adopted. During testing, input images are resized to $2,048\times 512$. Then, each dimension size is rounded up to the next multiple of $256$.~\tabref{tab:supp_seg_aug_circular_swinv2} describes the pre-processing and data augmentation pipelines.

\begin{table}[ht]
\begin{center}
\small
\begin{tabular}{c|c|c}
\specialrule{.15em}{.05em}{.05em} 
& Train Set & Test Set\\
\midrule
\makecell{Size Preprocessing} & \makecell{(i) Resize: $1,792 \times 448$\\ (ii) Random Crop: $448 \times 448$} & \makecell{(i) Resize: $1,792 \times 448$\\ (ii) Resize to next multiple of $224$}\\
\specialrule{.15em}{.05em}{.05em}
\makecell{Augmentation} & \makecell{(i) Random Horizontal Flipping \\ (ii) Photometric Distortion\\ (iii) Normalization} & \makecell{Normalization}\\
\specialrule{.15em}{.05em}{.05em}
\end{tabular}
\end{center}
\vspace{-0.5 cm}
\caption{\label{tab:supp_seg_aug_circular_swin} \textbf{Swin $+$ UperNet Semantic Segmentation preprocessing (circular shifts).}  Data preprocessing and augmentation used to train and test the default (Swin+UperNet) and adaptive (A-Swin+UperNet) models.}
\end{table}
\begin{table}[ht]
\begin{center}
\small
\begin{tabular}{c|c|c}
\specialrule{.15em}{.05em}{.05em} 
& Train Set & Test Set\\
\midrule
\makecell{Size Preprocessing} & \makecell{(i) Resize: $2,048 \times 512$\\ (ii) Random Crop: $512 \times 512$} & \makecell{(i) Resize: $2,042 \times 512$\\ (ii) Resize to next multiple of $256$}\\
\specialrule{.15em}{.05em}{.05em}
\makecell{Augmentation} & \makecell{(i) Random Horizontal Flipping \\ (ii) Photometric Distortion\\ (iii) Normalization} & \makecell{Normalization}\\
\specialrule{.15em}{.05em}{.05em}
\end{tabular}
\end{center}
\vspace{-0.5 cm}
\caption{\label{tab:supp_seg_aug_circular_swinv2} \textbf{SwinV2 $+$ UperNet Semantic Segmentation preprocessing (circular shifts).}  Data preprocessing and augmentation used to train and test the default (SwinV2+UperNet) and adaptive (A-SwinV2+UperNet) models.}
\end{table}

{\bf\noindent Data pre-processing (Standard Shifts).}
For the Swin+UperNet baseline and adaptive models, the default pre-processing pipeline from the official MMseg implementation~\cite{mmseg2020} was used. The same pipeline was used to evaluate the SwinV2+UperNet baseline and adaptive models. The pre-processing pipeline is detailed in~\tabref{tab:supp_seg_aug_standard}.

\begin{table}[t]
\begin{center}
\small
\begin{tabular}{c|c|c}
\specialrule{.15em}{.05em}{.05em} 
& Train Set & Test Set\\
\midrule
\makecell{Size Preprocessing} & \makecell{(i) Resize: $2,048 \times 512$\\ (ii) Random Crop: $512 \times 512$} & \makecell{Resize: $2,048 \times 512$}\\
\specialrule{.15em}{.05em}{.05em}
\makecell{Augmentation} & \makecell{(i) Random Horizontal Flipping \\ (ii) Photometric Distortion\\ (iii) Normalization} & \makecell{Normalization}\\
\specialrule{.15em}{.05em}{.05em}
\end{tabular}
\end{center}
\vspace{-0.5 cm}
\caption{\label{tab:supp_seg_aug_standard} \textbf{Semantic Segmentation preprocessing (standard shifts).} Data preprocessing and augmentation used to train and test the default and adaptive versions of both SwinV2+UperNet and A-SwinV2+UperNet models.}
\end{table}

{\bf\noindent Computational settings.}
ADE20K semantic segmentation experiments using Swin and SwinV2 models, both adaptive and baseline architectures, were trained on four NVIDIA A100 GPUs with an effective batch size of $16$ images for $160,000$ iterations. These settings are consistent with the official MMSeg configuration for the Swin+UperNet model.

\section{Additional semantic segmentation results}\label{sec:supp_results}
\begin{figure}[t]
\centering
\colorbox{white}{\begin{minipage}[t]{0.125\textwidth}
\centering \textbf{\scalebox{0.65}{Image}}
\end{minipage}}\colorbox{white}{\begin{minipage}[t]{0.125\textwidth}
\centering \textbf{\scalebox{0.65}{Shifted Image}}
\end{minipage}}\colorbox{white}{\begin{minipage}[t]{0.25\textwidth}
\centering \textbf{\scalebox{0.65}{Swin $+$ UperNet Predictions}}
\end{minipage}}\colorbox{white}{\begin{minipage}[t]{0.25\textwidth}
\centering\textbf{\scalebox{0.65}{A-Swin $+$ UperNet Predictions (Ours)}}
\end{minipage}}

\vspace{-0.1 cm}
\includegraphics[width=0.8\textwidth]{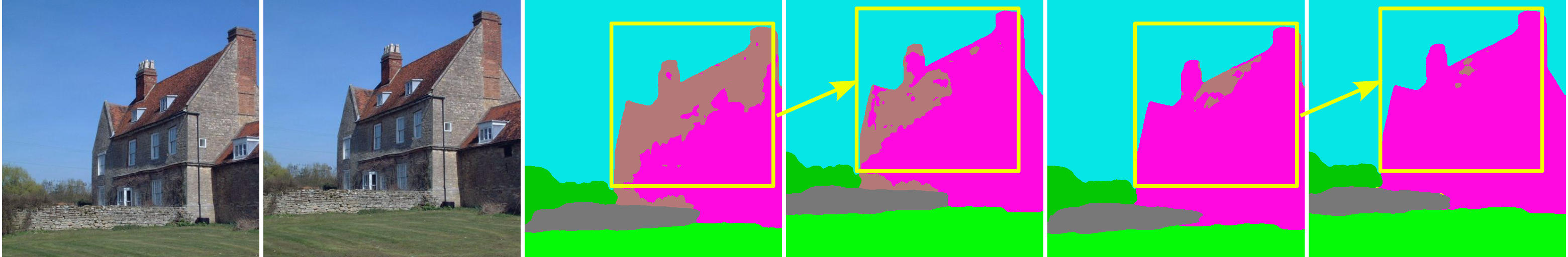}%

\includegraphics[width=0.8\textwidth]{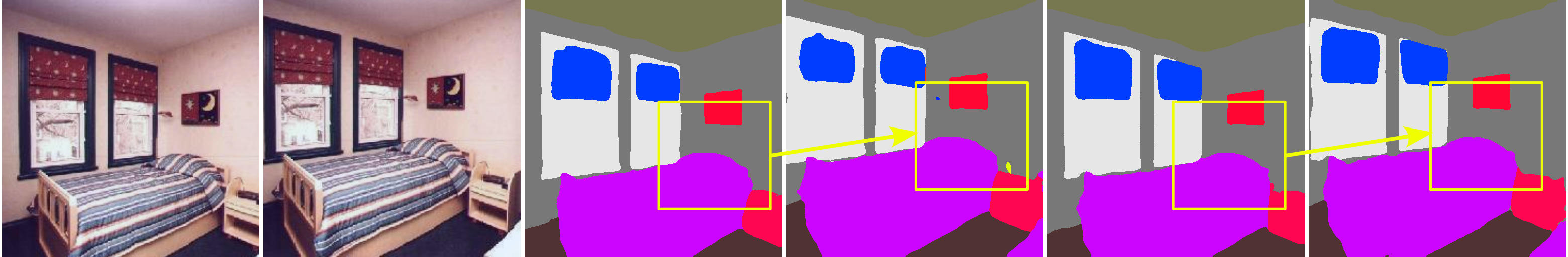}

\includegraphics[width=0.8\textwidth]{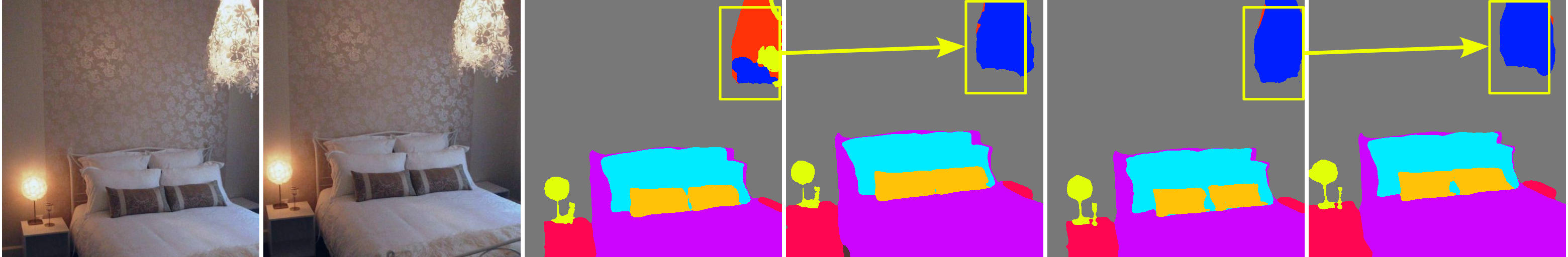}

\includegraphics[width=0.8\textwidth]{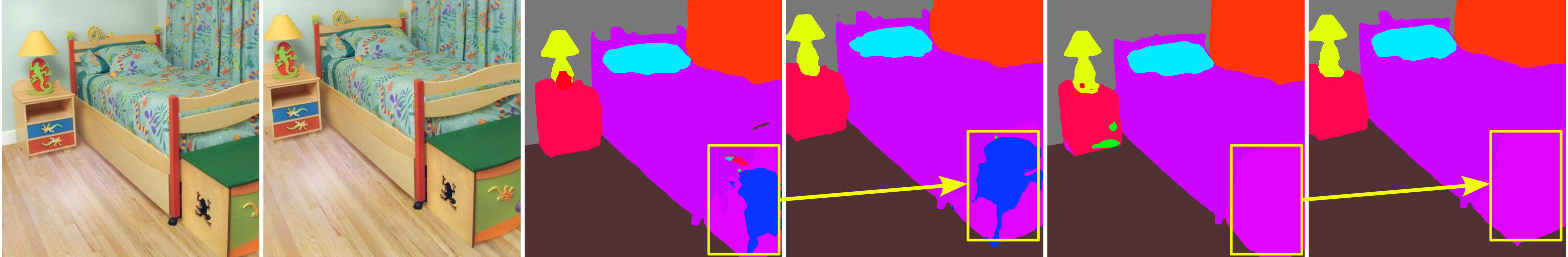}

\includegraphics[width=0.8\textwidth]{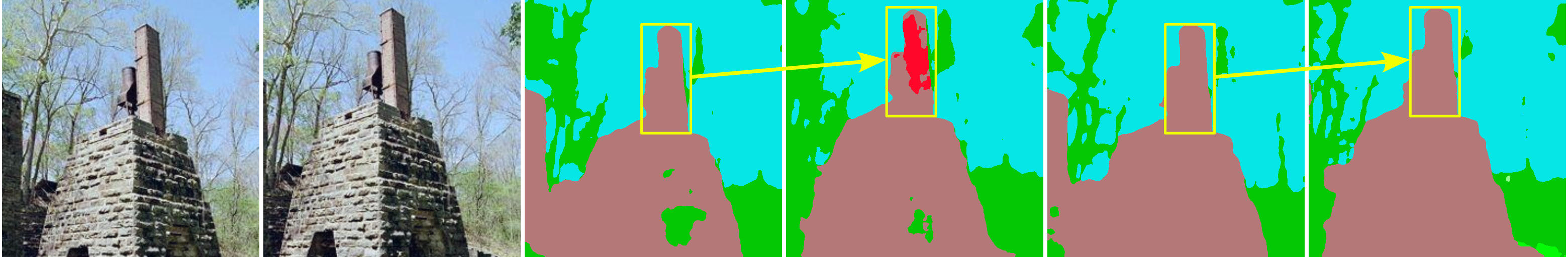}

\includegraphics[width=0.8\textwidth]{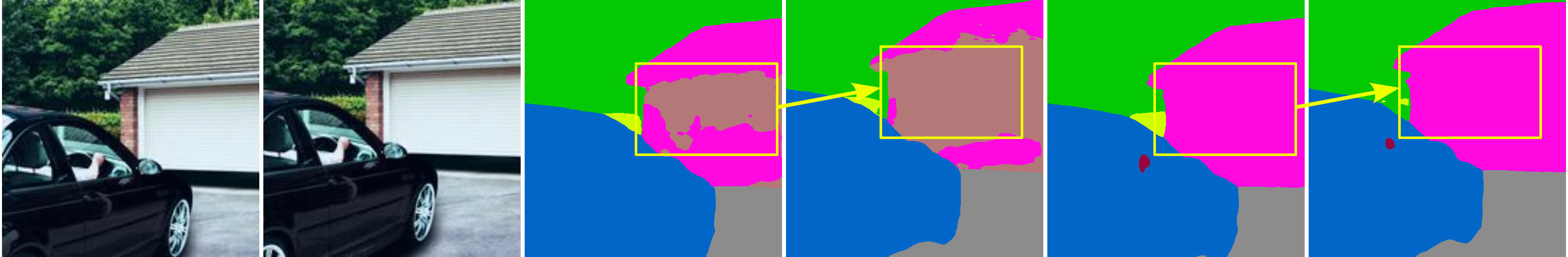}

\includegraphics[width=0.8\textwidth]{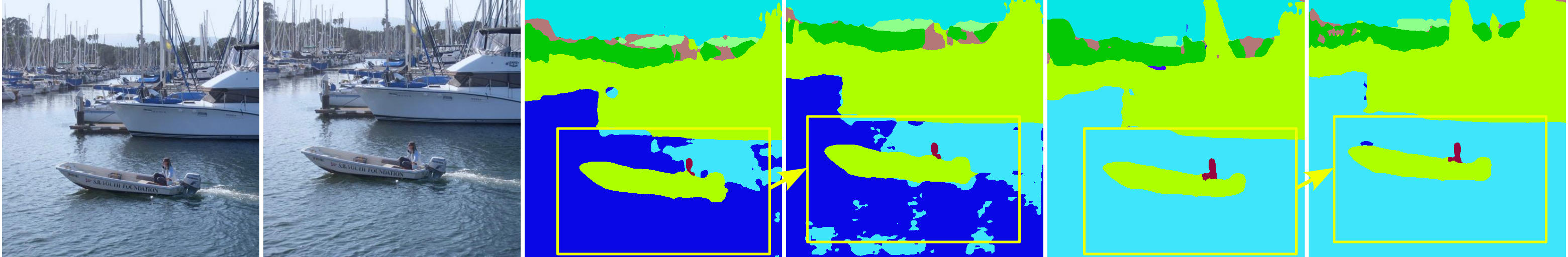}

\includegraphics[width=0.8\textwidth]{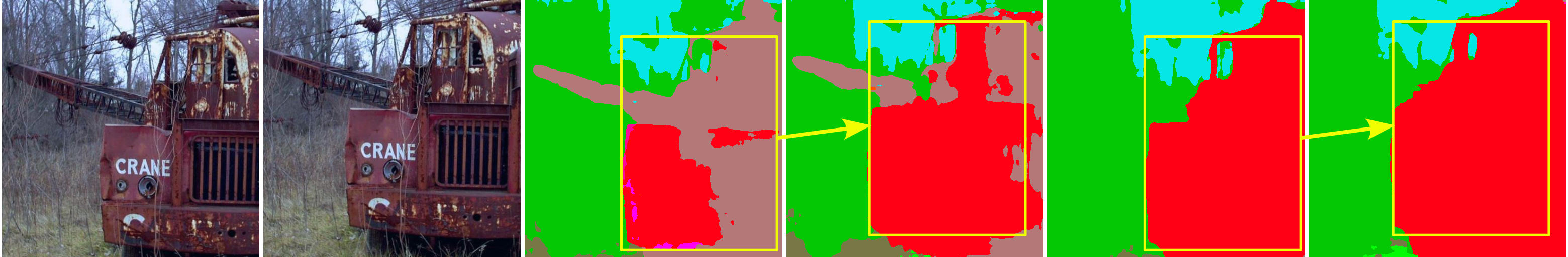}

\includegraphics[width=0.8\textwidth]{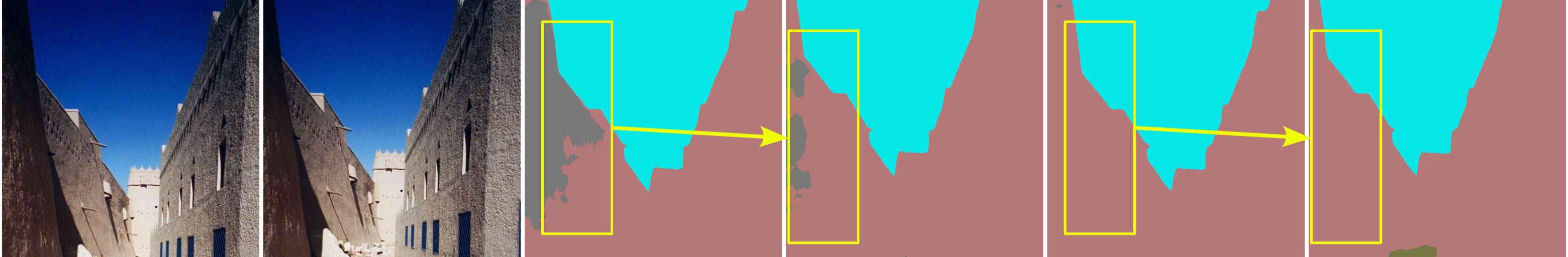}

\caption{\label{fig:supp_seg_qual_swin} \textbf{Swin + UperNet Semantic Segmentation under standard shifts:} Semantic segmentation results on the ADE20K dataset (standard shifts) via Swin backbones: Our A-Swin $+$ UperNet model is more robust to input shifts than the original Swin $+$ UperNet model, generating consistent predictions while improving accuracy. Examples of prediction changes due to input shifts are boxed in yellow.}
\vspace{-0.5cm}
\end{figure}

\begin{figure}[t]
\centering
\colorbox{white}{\begin{minipage}[t]{0.125\textwidth}
\centering \textbf{\scalebox{0.65}{Image}}
\end{minipage}}\colorbox{white}{\begin{minipage}[t]{0.125\textwidth}
\centering \textbf{\scalebox{0.65}{Shifted Image}}
\end{minipage}}\colorbox{white}{\begin{minipage}[t]{0.25\textwidth}
\centering \textbf{\scalebox{0.65}{SwinV2 $+$ UperNet Predictions}}
\end{minipage}}\colorbox{white}{\begin{minipage}[t]{0.25\textwidth}
\centering\textbf{\scalebox{0.65}{A-SwinV2 $+$ UperNet Predictions (Ours)}}
\end{minipage}}

\vspace{-0.1 cm}
\includegraphics[width=0.8\textwidth]{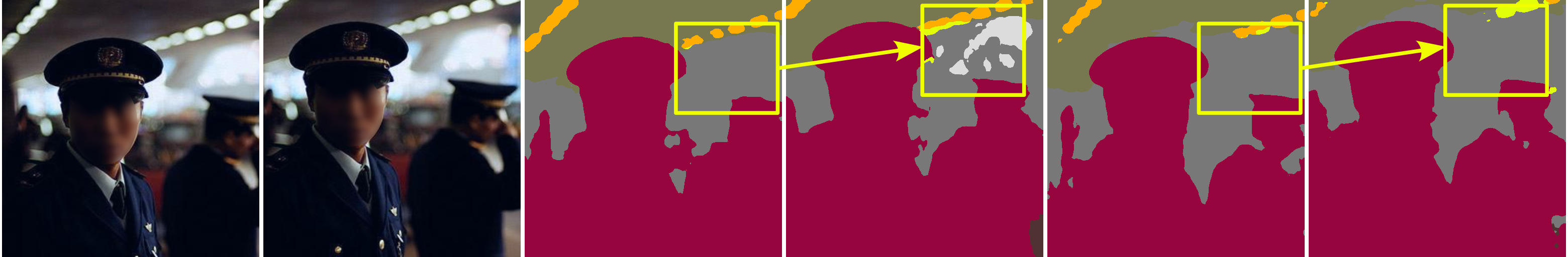}%

\includegraphics[width=0.8\textwidth]{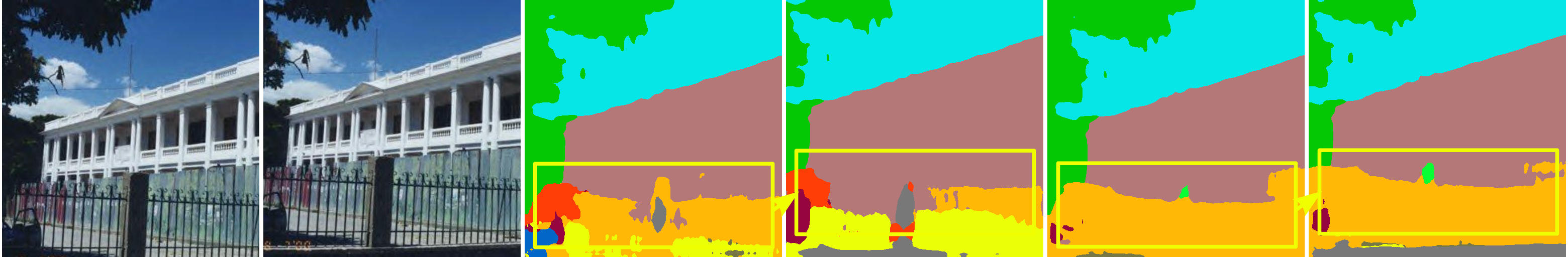}

\includegraphics[width=0.8\textwidth]{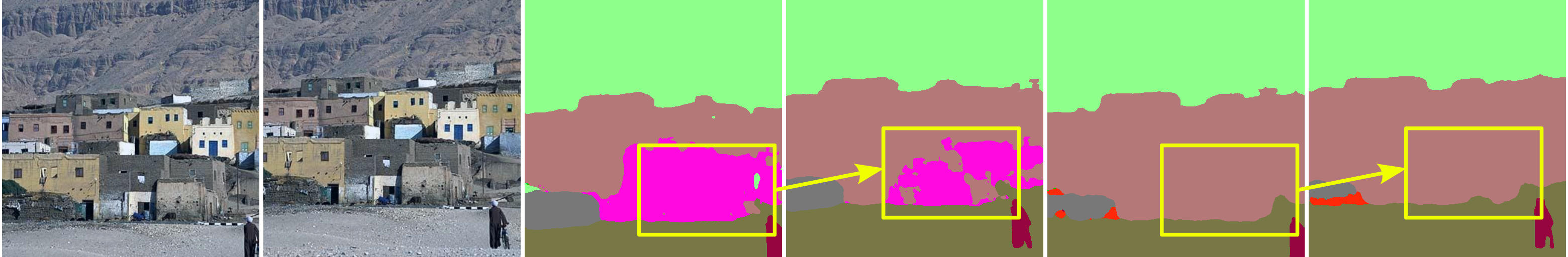}

\includegraphics[width=0.8\textwidth]{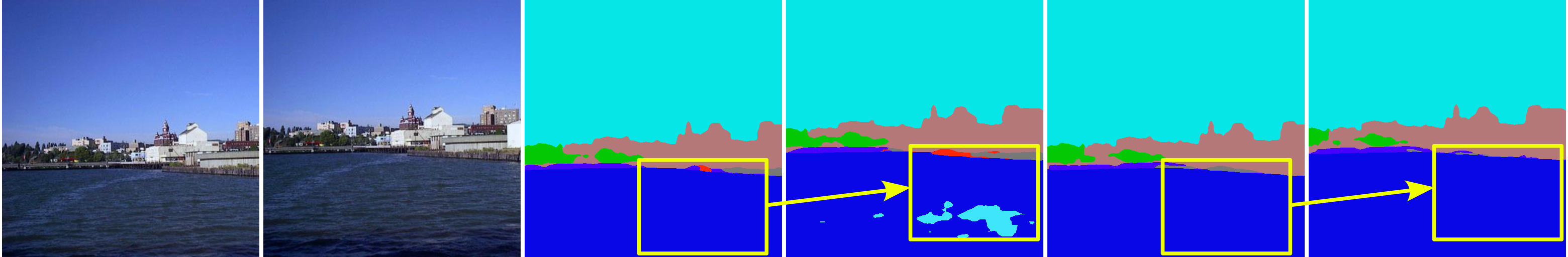}

\includegraphics[width=0.8\textwidth]{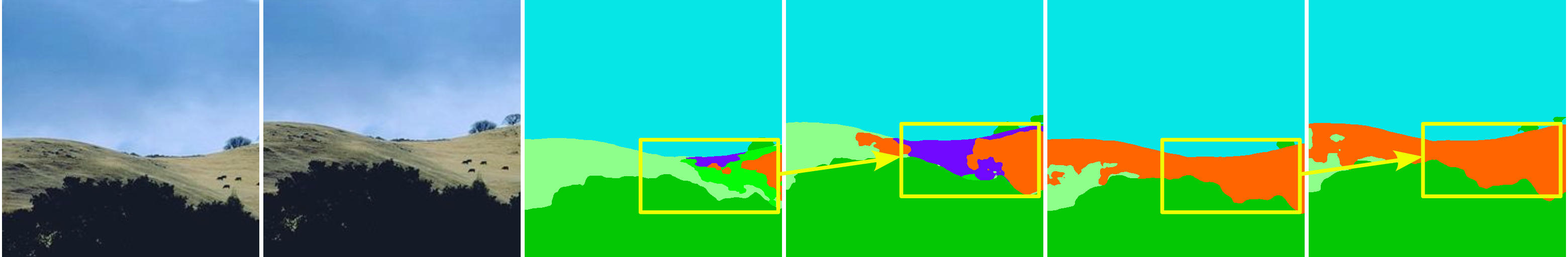}

\includegraphics[width=0.8\textwidth]{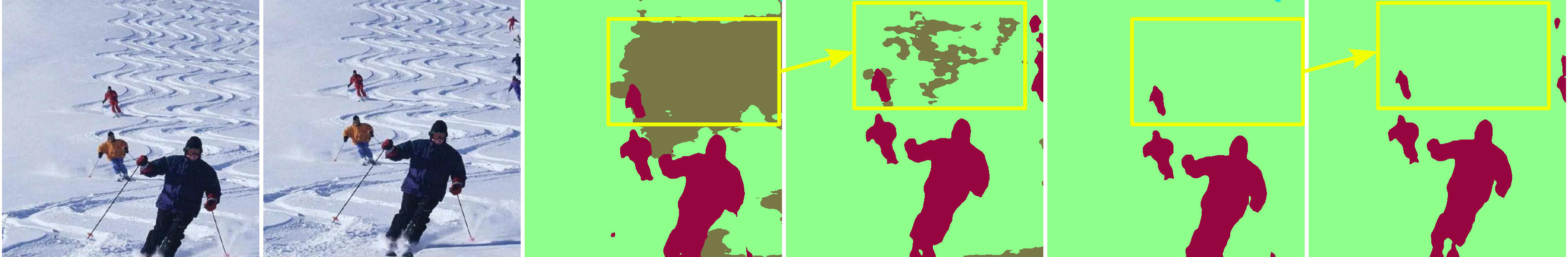}

\includegraphics[width=0.8\textwidth]{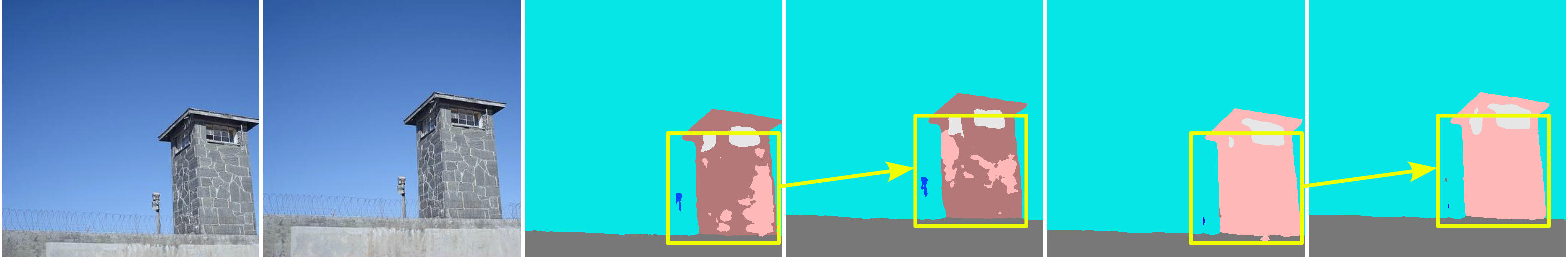}

\includegraphics[width=0.8\textwidth]{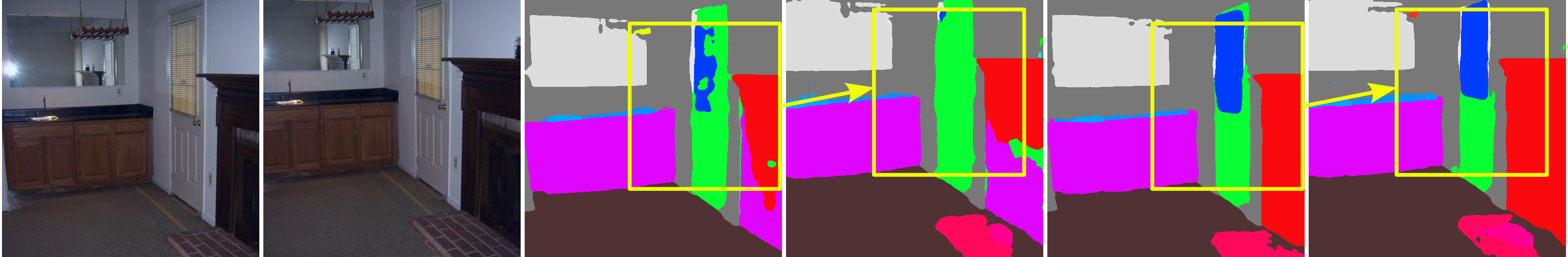}

\includegraphics[width=0.8\textwidth]{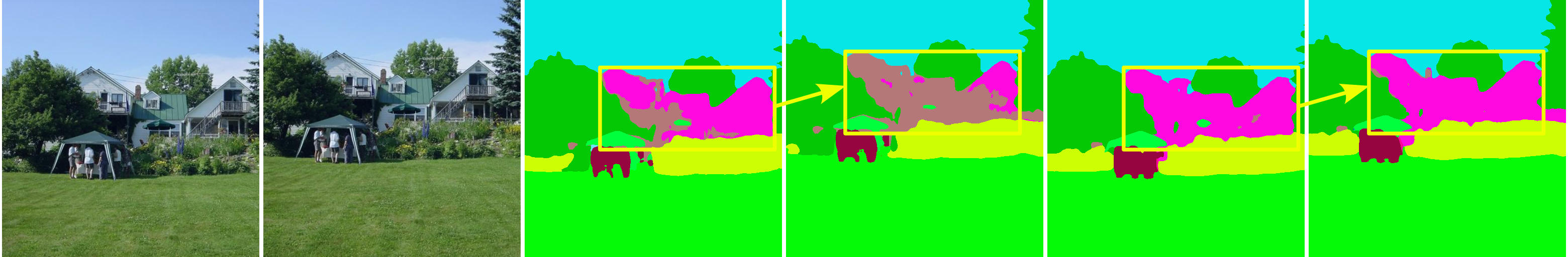}

\caption{\label{fig:supp_seg_qual_swinv2} \textbf{SwinV2 + UperNet Semantic Segmentation under standard shifts:} Additional semantic segmentation results on the ADE20K dataset (standard shifts) via SwinV2 backbones: Our adaptive model  improves both segmentation accuracy and shift consistency with respect to the original SwinV2 $+$ UperNet model. Examples of prediction changes due to input shifts are boxed in yellow.}
\vspace{-0.5cm}
\end{figure}

\figref{fig:supp_seg_qual_swin} and~\ref{fig:supp_seg_qual_swinv2} show examples of semantic segmentation predicted masks for our A-Swin+UperNet and A-SwinV2+Upernet models, respectively. Illustrations include masks obtained with their corresponding baselines, showing the improved robustness of our models against input shifts.

\section{Additional image classification experiments} \label{sec:supp_additional_experiments}
\subsection{Robustness to out of distribution images} \label{sec:supp_ood}

We evaluate the image classification performance of our adaptive models on out-of-distribution inputs. Precisely, we measure the robustness of our proposed A-Swin-T model to images with randomly erased patches and vertically flipped images.

\myparagraph{Experiment Setup.}
Following the evaluation protocol of \citet{chaman2021truly}, we compare the performance of three versions of Swin-T: (i) its default model, (ii) our proposed adaptive model (A-Swin-T), and (iii) the default Swin-T architecture trained on circularly shifted images, denoted as Swin-T DA (data augmented).

All three models are trained on CIFAR-10. Swin-T and A-Swin-T use the default data pre-processing used for CIFAR-10 (refer to \secref{sec:supp_exp_details}), while Swin-T DA uses the default pre-processing plus circular shifts with offsets uniformly selected between $-224$ and $224$ pixels. Note that CIFAR-10 samples are resized from $32 \times 32$ to $224 \times 224$ pixels during pre-processing. So, the circular offsets are effectively selected between $-32$ and $32$ pixels.

During testing, the patch size is randomly selected between 0 and $\text{max}$, where $\text{max} \in \{28, 42, 56, 70\}$ pixels. Since CIFAR-10 samples are rescaled to $224$ x $224$ pixels, the maximum patch sizes are effectively $\{4, 6, 8, 10\}$ pixels.

\myparagraph{Results.}
\tabref{tab:supp_ood_patches} shows the top-1 classification accuracy and circular shift consistency of the three Swin-T models of interest on images with randomly erased patches. Across patch sizes, while the default model and Swin-T DA decrease their shift consistency by at least 1.5$\%$, our A-Swin-T preserves its perfect shift consistency. Our adaptive model also gets the best classification accuracy in all cases, improving by more than $1\%$. This suggests that, despite not being explicitly trained on this transformation, our A-Swin model is more robust than the default and DA models, obtaining better accuracy and consistency across scenarios.

On the other hand, \tabref{tab:supp_ood_flips} shows the shift consistency and classification accuracy of the three Swin models evaluated under vertically flipped images. Even in such a challenging case and without any fine-tuning, our A-Swin-T model retains its perfect shift consistency, improving over the default and data-augmented models by at least $5\%$. Our adaptive model also outperforms the default and data-augmented models in terms of classification accuracy, both on flipped and unflipped images, by a significant margin.

\begin{table}[t]
\centering
\small
\rowcolors{1}{bg_blue}{}
\resizebox{\columnwidth}{!}{%
\begin{tabular}{c | c c | c c | c c | c c | c c}
\specialrule{.15em}{.05em}{.05em}
\hiderowcolors 
Model & \multicolumn{2}{c|}{max $=0$} & \multicolumn{2}{c|}{max $=28$} & \multicolumn{2}{c|}{max $=42$} & \multicolumn{2}{c|}{max $=56$} & \multicolumn{2}{c}{max $=70$}\\
 & Acc. & C-Cons. & Top-1 Acc. & C-Cons. & Top-1 Acc. & C-Cons. & Top-1 Acc. & C-Cons. & Top-1 Acc. & C-Cons.\\
 \midrule
  Swin-T (Default)    & $90.21$ & $82.69$ & $90.16$ & $81.75$ & $89.71$ & $81.42$ & $89.02$ & $80.4$ & $87.83$ & $79.58$ \\
  Swin-T DA    & $92.32$ & $94.3$ & $91.81$ & $94.19$ & $91.6$ & $94.16$ & $90.42$ & $93.01$ & $89.23$ & $92.73$ \\
  \showrowcolors
  A-Swin-T \textbf{(Ours)}   & $\bf 93.53$ & $\bf 100$ & $\bf 93.27$ & $\bf 100$ & $\bf 92.71$ & $\bf 100$ & $\bf 91.57$ & $\bf 100$ & $\bf 90.3$ & $\bf 100$ \\

\specialrule{.15em}{.05em}{.05em}
\end{tabular}%
}
\vspace{-0.25 cm}
\caption{\textbf{Performance on images with randomly erased patches.} Top-1 classification accuracy $(\%)$ and shift consistency $(\%)$ on CIFAR-10 test images with randomly erased square patches. Max corresponds to the largest possible patch size, as sampled from a uniform distribution $\mathcal{U}\{0,\text{max}\}$.}
\label{tab:supp_ood_patches}
\end{table}

\begin{table}[t]
\centering
\small
\vspace{-0.15cm}
\rowcolors{1}{bg_blue}{}
\begin{tabular}{c | c c | c c}
\specialrule{.15em}{.05em}{.05em}
\hiderowcolors
Model & \multicolumn{2}{c|}{Unflipped} & \multicolumn{2}{c}{Flipped}\\
 & Top-1 Acc.$(\%)$ & C-Cons.$(\%)$ & Top-1 Acc.$(\%)$ & C-Cons.$(\%)$\\
 \midrule
  Swin-T (Default)    & $90.21$ & $82.69$ & $50.41$ & $82.24$\\
  \showrowcolors
  Swin-T DA    & $92.32$ & $94.3$ & $51.74$ & $94.67$\\
  A-Swin-T \textbf{(Ours)}   & $\bf 93.53$ & $\bf 100$ & $\bf 52.07$ & $\bf 100$\\

\specialrule{.15em}{.05em}{.05em}
\end{tabular}
\vspace{-0.25 cm}
\caption{\textbf{Performance on flipped images.} Top-1 classification accuracy and shift consistency on vertically flipped CIFAR-10 test images.}
\label{tab:supp_ood_flips}
\end{table}

\subsection{Sensitivity to input shifts} \label{sec:supp_sensitivity}
To measure the shift consistency improvement of our proposed adaptive models over the default ones, we conduct a fine-grained evaluation of shift consistency by analyzing different offset magnitudes.

\myparagraph{Experiment setup.} We test the shift sensitivity of our four proposed adaptive ViTs on CIFAR-10. Circular shift consistency is evaluated on images shifted by an offset randomly selected from four different ranges: $[0, 56]$, $[0, 112]$, $[0, 168]$, and $[0, 224]$ pixels. Note that CIFAR-10 images are resized from $32 \times 32$ to $224 \times 224$ pixels. So, the effective intervals correspond to $[0, 8]$, $[0, 16]$, $[0, 24]$, and $[0, 32]$ pixels, respectively. By gradually increasing the offset magnitude, we can understand the advantages of our adaptive models over their default versions in a more general manner.

\myparagraph{Results.} \tabref{tab:supp_sensitivity} shows the circular shift consistency of all four models under different shift magnitudes. While the default versions monotonically decrease their shift consistency with respect to the offset magnitude, our method shows a perfect shift consistency across scenarios.

Despite the strong consistency obtained by default ViTs via data augmentation, our adaptive models outperform them by more than $9\%$ without any fine-tuning. This shows the benefits of our adaptive models, regardless of the shift magnitude.

\begin{table}[t]
\centering
\small
\rowcolors{1}{bg_blue}{}
\begin{tabular}{c | c | c | c | c}
\specialrule{.15em}{.05em}{.05em}
\hiderowcolors
Model & Offset $\in \{0,\dots,8\}$ & Offset $\in \{0,\dots,16\}$ & Offset $\in \{0,\dots,24\}$ & Offset $\in \{0,\dots,32\}$\\
 \midrule
  Swin-T & $92.00\pm.23$ & $89.65\pm.9$ & $88.93\pm.09$ & $88.13\pm.14$ \\
  \showrowcolors
  A-Swin-T \textbf{(Ours)}   & $\bf 100$ & $\bf 100$ & $\bf 100$ & $\bf 100$ \\
  SwinV2-T & $92.13\pm.04$ & $90.43\pm.17$ & $89.67\pm.08$ & $88.75\pm.15$ \\
  A-SwinV2-T \textbf{(Ours)} & $\bf 100$ & $\bf 100$ & $\bf 100$ & $\bf 100$ \\
  CvT-13 & $88.99\pm.1$ & $87.42\pm0.16$ & $86.96\pm.1$ & $86.84\pm.06$ \\
  A-CvT-13 \textbf{(Ours)}    & $\bf 100$ & $\bf 100$ & $\bf 100$ & $\bf 100$ \\
  MViTv2-T & $91.49\pm.04$ & $90.57.\pm.11$ & $90.46\pm.08$ & $90.22\pm.1$ \\
  A-MViTv2-T \textbf{(Ours)}   & $\bf 100$ & $\bf 100$ & $\bf 100$ & $\bf 100$ \\

\specialrule{.15em}{.05em}{.05em}
\end{tabular}%
\vspace{-0.25 cm}
\caption{\textbf{Consistency under different shift magnitudes.} Shift consistency $(\%)$ of our adaptive ViT models under small, medium, large, and very large shifts. Models trained and evaluated on CIFAR-10 under a circular shift assumption.}
\label{tab:supp_sensitivity}
\end{table}

\subsection{Replacing adaptive modules in pre-trained ViTs}
\label{sec:supp_pretrained}

We ran additional experiments plugging in our proposed adaptive modules on pre-trained default ViTs. Note that, while replacing default ViT modules with adaptive ones guarantees a truly shift-equivariant model ($100\%$ shift consistency), it is expected for the classification accuracy to decrease since the pre-trained weights have not been trained to adaptively select different token or window representations. Nevertheless, we are interested in exploring this scenario as a refined initialization strategy, in order to evaluate its effect in terms of classification accuracy and shift consistency.

\myparagraph{Experiment setup.}
Given a default CvT-13 model trained on CIFAR-10, we replace its components with our proposed framework. We denote this model as \text{CvT-13 $+$ Adapt}. Next, we fine-tune the model for $10$ and $20$ epochs and evaluate its top-1 classification accuracy as well as shift consistency.

\myparagraph{Results.}
\tabref{tab:supp_pretrained} shows the top-1 classification accuracy and shift consistency of CvT-13 $+$ Adapt, as well as the default CvT-13 and our proposed A-CvT-13 model. Without any fine-tuning, CvT-13 + Adapt attains perfect shift consistency just by plugging in our proposed adaptive modules. However, its top-1 classification accuracy decreases by more than $30\%$ with respect to the default model. By fine-tuning the model for $10$ and $20$ epochs, its classification accuracy boosts up to $91.72\%$ and $92.35\%$, respectively. This implies that CvT-13 + Adapt is capable of outperforming the default model both in consistency and accuracy by plugging in our adaptive components plus a few fine-tuning iterations.

On the other hand, while promising results are obtained via fine-tuning, our adaptive model A-CvT-13, fully trained for $90$ epochs using random initialization and the default training settings, still attains the best classification accuracy and shift consistency results.

\begin{table}[t]
\centering
\small
\rowcolors{1}{}{bg_blue}
\begin{tabular}{c|c|c}
\specialrule{.15em}{.05em}{.05em}
\hiderowcolors
Model & Top-1 Acc.$(\%)$ & C-Cons.$(\%)$\\
 \midrule
  CvT-13 (Default)    & $90.12$ & $76.54$ \\
  CvT-13 $+$ Adapt (No Fine-tuning)    & $57.4$ & $100$ \\
  CvT-13 $+$ Adapt ($10$ epoch Fine-tuning)    & $91.72$ & $100$ \\
  \showrowcolors
  CvT-13 $+$ Adapt ($20$ epoch Fine-tuning)    & $92.35$ & $100$ \\
  A-CvT-13 \textbf{(Ours)}   & $\bf 93.87$ & $\bf 100$ \\

\specialrule{.15em}{.05em}{.05em}
\end{tabular}
\caption{\textbf{Incorporating shift-equivariant modules on pre-trained ViTs.} Our shift-equivariant ViT framework allows plugging-in shift equivariant modules on pre-trained models (e.g. CvT-13), improving on classification accuracy after a few fine-tuning iterations while preserving its perfect shift consistency. Results shown on CvT-13 trained on CIFAR-10 under a circular shift assumption.}
\label{tab:supp_pretrained}
\end{table}

\end{document}